\documentclass[5p,time]{elsarticle}

\usepackage{lineno,hyperref}
\modulolinenumbers[5]

\usepackage{moreverb,url}
\usepackage{color}
\usepackage{graphicx}
\usepackage{epsfig} 
\usepackage{mathptmx} 
\usepackage{times} 
\usepackage{amsmath} 
\usepackage{amssymb}  
\usepackage{epstopdf}
\usepackage{algorithm}
\usepackage{algpseudocode}
\usepackage{array}
\usepackage[caption=false,font=footnotesize]{subfig}
\usepackage{url}
\usepackage{multirow}
\usepackage{enumerate}
\graphicspath{
	{figs/}
}

\journal{Elsevier Robotics and Autonomous Systems}

\begin{document}
\sloppy

\begin{frontmatter}

\title{Efficient and Robust Pedestrian Detection \\ using Deep Learning for Human-Aware Navigation
}

\author{Andr\'{e} Mateus\textsuperscript{1}, David Ribeiro\textsuperscript{1}, Pedro Miraldo\textsuperscript{1,2}, and Jacinto C. Nascimento\textsuperscript{1}}
\address{\textsuperscript{1}Instituto de Sistemas e Rob\'{o}tica (LARSyS), Instituto Superior T\'{e}cnico, Lisboa, Portugal.\\
\textsuperscript{2}Department of Automatic Control, KTH Royal Institute of Technology, Stockholm, Sweden.\\ Corresponding Author: {\tt andre.mateus@tecnico.ulisboa.pt}}

\begin{abstract}
This paper addresses the problem of Human-Aware Navigation (HAN), using multi camera sensors to implement a vision-based person tracking system. The main contributions of this paper are as follows: a novel and efficient Deep Learning person detection and a standardization of human-aware constraints. In the first stage of the approach, we propose to cascade the Aggregate Channel Features (ACF) detector with a deep Convolutional Neural Network (CNN) to achieve fast and accurate Pedestrian Detection (PD). Regarding the human awareness (that can be defined as constraints associated with the robot's motion), we use a mixture of asymmetric Gaussian functions, to define the cost functions associated to each constraint. Both methods proposed herein are evaluated individually to measure the impact of each of the components. The final solution (including both the proposed pedestrian detection and the human-aware constraints) is tested in a typical domestic indoor scenario, in four distinct experiments. The results show that the robot is able to cope with human-aware constraints, defined after common proxemics and social rules.
\end{abstract}


\end{frontmatter}

\section{Introduction}
For robots to interact naturally with humans in their social
environments, they must have the ability to plan their motion
accounting for typical social norms. In this paper, we address
the robot navigation in the presence of humans, resorting to multi cameras (static outside and/or onboard
cameras) for the vision-based person tracking system.

One of the research focus in robotics is Human-Robot Interaction and its role in social environments \cite{article:kruse:2013}. When people
think of a robot interacting with a person, what comes to mind is a
robot that can speak with her or hand over some object. However, the
motion itself is of great importance in a social context (e.g., when
a robot is requested to fetch an item), or simply when a normal
navigation behavior needs to be adjusted according to proxemics
rules, so it does not disturb people. The study of robot navigation
in the presence of people is called Human-Aware Navigation (HAN).

Most approaches to HAN in the literature use only sensors onboard
the robot \cite{article:Sisbot:2006, 34,21,article:pandey:2010}. Even though those sensors bring the advantage of
context-independence, it is useful to have other external sensors,
which can add more information about the environment, not only in
terms of coverage space, but also in terms of precision on the
estimation of the person's position. Hence, in this paper, in
addition to the onboard camera, external cameras were mounted on
the ceiling and used for the pedestrian detection (PD) task. This
setup ensures a broader perception of the environment, capable of
seeing/detecting both humans and robots at the same time.

Furthermore, in this paper we address the importance of
incorporating deep learning (DL) in an HAN based architecture. The
integration of DL provides both: (i) {\em efficiency} and (ii) {\em
	robustness} to the pedestrian detection task, as detailed next.

Traditionally, the detection task is usually accomplished through
the sliding window paradigm, based on a exhaustive search over the
image to find the object positions (e.g.,
\cite{DalalCVPR2005,FelzenszwalbPAMI2010,ViolaIJCV2004,ZhuCVPR2010}).
During this search, features for each window location are computed (possibly at multiple scales), and later evaluated by a
classifier. However, such procedure can easily become intractable
due to the substantial number of windows and the complexity of the
features processing. Thus, for the use of richer and more expensive
features, it is necessary to constrain the computations to a
restricted number of locations, considering only certain regions of
interest.

Accordingly, we propose a cascade of the ACF detector
\cite{DollarPAMI2014} with a CNN for PD. The ACF allows to obtain a
{\em selective search} that identifies promising image regions for
the presence of pedestrians (i.e., proposals). This alleviates the
CNN task, since the number of windows (i.e., proposals) to classify
is now substantially reduced. The advantages of this cascade are
twofold. Firstly, we only use the more expensive computations (i.e.,
the CNN classification) in the promising ACF proposals.
Consequently, operating under this regions of interest paradigm
\cite{Girshick2014,Hosang2015Cvpr}, allows to speed up the detection
procedure and perform pedestrian detection at real time
requirements. The methodology previously described, contributes to
the {\em efficiency} goal. Secondly, the first stage of the cascade
(i.e., the ACF) provides a large number of false positives (FP). This
number of FPs is drastically reduced by the application of the CNN,
while maintaining most of the true positives (TP). This means that
the errors corresponding to these FPs provided by the ACF are solved
within the CNN processing. This approach allows to achieve the {\em robustness}
goal.

The CNN model results from first, pre-training with a large object
dataset \cite{RussakovskyIJCV2015}, and then fine-tuning
(i.e., re-training) with a PD dataset \cite{DalalCVPR2005}. This transfer
learning procedure \cite{YosinskiNIPS2014} is adopted because it
improves the final model. In order to guarantee independence from the
particular application in the HAN context, no further fine-tuning is
performed (although the CNN could be fine-tuned again with a
specific HAN dataset).

Finally, we also propose a novel solution for HAN resorting to
multiple cameras (onboard and offboard), for people state
estimation, coupled with the aforementioned DL strategy. A costmap
is computed by combining several constraints associated with HAN.
Each time the robot receives a new goal, it computes a path on that
costmap.
When compared with state-of-the-art approaches, the main
contributions presented in this paper are: $(i)$ a novel and efficient
technique for people detection; and $(ii)$ a standardization of human-aware
constraints. Moreover, since all the computations are performed using CPU, this methodology is an economically
viable solution, that reaches the intended runtime figures and accuracy requirements. Therefore, the mentioned approach
can be integrated in robots that only have onboard CPUs (no GPUs).

The solution is tested in simulated and in realistic scenarios. The results show that the proposed solution fulfils the
aforementioned goals.

This paper proposes novel extensions of the authors' previous works, namely \cite{mateus:2015,ribeiro2017ICARSC}.
More specifically, in \cite{mateus:2015} HAN constraints are introduced, and in \cite{ribeiro2017ICARSC} a real-time PD algorithm is proposed. Comparing the present paper with \cite{ribeiro2017ICARSC}, additional outside cameras are used, more HAN constraints are introduced and tested, and additional details are included (e.g. in the method's evaluation and the related work).

\subsection{State-of-the-Art}

To give a better comprehensive understanding of the contributions presented herein, we next review the related work on both PD and HAN, respectively.

\subsubsection{Pedestrian Detection:} One of the main goals in our framework is to achieve an accurate and fast Pedestrian Detection algorithm. This has been one of the major topics addressed in the computer vision community, surveys are available in \cite{DollarPAMI2012,BenensonECCV2014}.

Classically, the PD problem has been addressed by using conventional handcrafted features (e.g., image gradient, HOG, wavelets, etc) that have plateaued in recent years. Some of the popular handcrafted methods for PD include: the Aggregate Channel Features (ACF) \cite{DollarPAMI2014}, where individual pixel lookups are extracted from the concatenation of the LUV, histogram of oriented gradients and gradient magnitude image channels. These pixels serve as features to be applied in a boosted decision trees classifier. The Locally Decorrelated Channel Features (LDCF) \cite{NamNIPS2014} is a more accurate variant of ACF (but also slower), which decorrelates the previously mentioned image channels resorting to linear filters. This idea of adding a filtering step for the features, was further studied in the work of \cite{Zhang2015Cvpr}.

Nevertheless, the application of deep compositional architectures, namely, Convolutional Neural Networks, to the tasks of image classification, localization and detection \cite{RussakovskyIJCV2015}, have significantly boosted the state-of-the-art.
As such, the application of deep learning to PD arises as a natural forthcoming step.
Indeed, deep learning based architectures learn hierarchical features \cite{Goodfellow-et-al-2016-Book} that make it possible to reach a better classification performance than using the handcrafted ones.

A dataset of substantial size and containing the corresponding annotations, contributes to reduce overfitting during the training of CNN models (considering their significant number of parameters) \cite{KrizhevskyNIPS2012}.
This is, however, an apparent limitation since this problem can be addressed by transferring parameters from an already trained CNN model (with datasets belonging to other tasks, for example, generic object classification) to the  model of interest. This CNN concerning the model of interest, is then re-trained (i.e., fine-tuned) with the dataset corresponding to the specific problem.

The computations associated with the CNN are expensive when compared to the ones required by methods using handcrafted features. Therefore, to improve the detector's speed, an hybrid solution can be adopted by cascading a faster and shallower method, based on handcrafted features, with a deep CNN. The handcrafted approach generates proposals (i.e., promising regions for the pedestrians locations), whose classification is refined by the CNN (i.e., the accuracy is enhanced by removing false positives).

Currently, various PD deep learning approaches have been developed in the literature, mainly by extending the main pipeline of successful object detectors, such as: R-CNN \cite{Girshick2014}, Fast R-CNN \cite{girshick15fastrcnn} and Faster R-CNN \cite{ren2015faster}. This pipeline comprises a combination of: $(i)$ proposal extraction and $(ii)$ CNN evaluation (being identical to the hybrid scheme, but more generic). For example, in \cite{Li2015}, the ACF is used in conjunction with the Fast R-CNN architecture, which has two CNN branches for small and large scales. The work of \cite{Cai2016} is able to improve the proposal extraction module of the Faster R-CNN scheme, resorting to diverse network outputs in order to detect objects of different scales. The work presented in \cite{Zhang2016} uses the region proposal network (CNN based) from Faster R-CNN and refines the detections resorting to CNN features and boosted decision trees.

In our approach, we adopt a similar architecture to \cite{HosangCVPR2015,RibeiroPR2016}, which follows the R-CNN pipeline. However, we do not apply any bounding box regression, contrasting to \cite{HosangCVPR2015}. Besides, and since we are concerned with speed, we introduce an ACF score rejection threshold. Below this value, the proposals are eliminated and are not processed by the CNN, allowing to achieve faster running time figures. Our main focus is devoted to the integration of the Computer Vision module (PD method) with the HAN module, in order to achieve an accurate and fast enough system.

\subsubsection{Human-Aware Navigation:} Regarding the HAN, a planner approach can be found in \cite{article:Sisbot:2006}. This work  focuses on human comfort, which is addressed by three criteria: preventing personal space invasions; navigating in the humans' field of view (FOV); and preventing sudden appearances in the FOV of humans. Those criteria are modeled as cost functions in a 2D costmap and path planning is performed with $A^{\star}$ algorithm. Even though the HAN planner accounts for replanning if people move, it does not adapt their personal space during the motion. With that in mind, two extensions to HAN are proposed:
\begin{itemize}
	\item{A prediction cost function which, by increasing the cost in front of a moving human, decreases the probability of the robot entering that area \cite{article:kruse:2010}}; and
	\item{The concept of compatible paths, which means that two paths are compatible if both agents can follow their paths (reaching the goal position), without any deadlocks \cite{article:kruse:2012}.}
\end{itemize}

An alternative approach was proposed by \cite{34}, which differs from HAN planner on the considered constraints and their formulation. Instead of focusing simply on human comfort, constraints concerning social rules (e.g., navigate on the right side of narrow passages) and low-level human navigation behavior (e.g., face direction of movement) are also taken into account. Another important issue related with human comfort, in a social context, is the interference with humans interacting with other humans and/or objects. This issue is tackled by \cite{21} where, besides considering proxemics and the back space of a person, a constraint is included to model the space between interacting entities. Other important work was presented in \cite{article:pandey:2010}. The authors presented a framework for planning a smooth path through a set of milestones. Those are added, deleted, and/or modified, based on the static and dynamic components of the environment.

More recently, \cite{article:kruse:2013} defined the three goals for Human-Aware Navigation: human comfort (e.g., space that people keep from each other in different contexts, known as the theory of proxemics \cite{33}, and velocity that robots navigate close to humans \cite{40}); respect social rules; and mimic low-level human behavior.

\subsection{Outline of the Paper}

This paper is organized as follows. Section \ref{sec:localization} introduces the main stages of the proposed vision-based framework. Section ~\ref{sec:Methodology-for-PD} describes the PD methodology. Section \ref{sec:Material-and-methods} is related with how the deep learning methodology is integrated in the navigation setup. To accomplish this, the CNN training must be performed (Section~\ref{sec:deep-trn-tst}), as well as the adaptation of the CNN (Section~\ref{sec:Adaptation of the CNN}). Experimental evaluation is conducted in Section~\ref{sec:Experimental-results}, in which, we evaluate the performance of the proposed PD methodology in the INRIA dataset (Section \ref{sec:Results-CNN-INRIA}) and in two real scenarios comprising the ``corridor'' and ``MBOT'' sequences (Section \ref{sec:Results-CNN-Corridor-MBOT}). Section \ref{sec:used-constraints} presents the HAN constraints that will be used in the proposed framework. In Section~\ref{sec:Simulation-Experiments}, we evaluate the HAN constraints, and in Section~\ref{sec:results-simul-envir}, we present the results with the complete framework (PD + HAN). Section~\ref{sec:conclusions} concludes the paper. The appendix provides additional details and formalizations, such as, the description of the CNN and the operations used, in \ref{sec:CNN-model}, and the runtime comparison between exhaustive search and the cascade ACF+CNN (including the threshold operation), in \ref{sec:comp_cascades_exhaustive_search}.

\begin{figure*}[t]
	\begin{center}
		\includegraphics[width=0.95\textwidth]{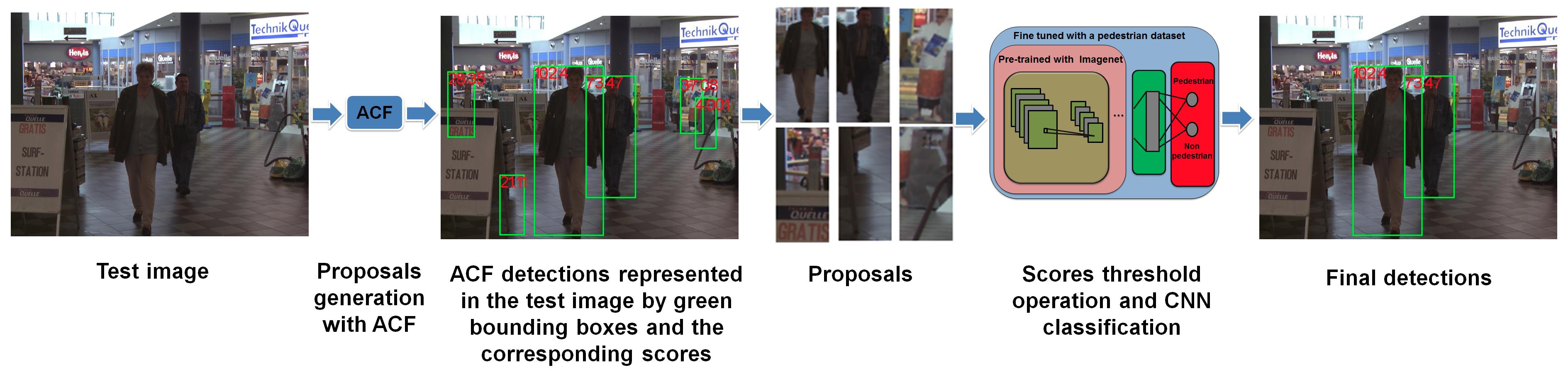}
	\end{center}
	\caption{Illustration of the proposed methodology, cascading the ACF (non-deep) detector and the deep CNN. First, the ACF detector performs selective search identifying promising image regions that might contain pedestrians, i.e., generates pedestrian proposals (see the green rectangles in the third image of the figure). The candidate proposals, in the RGB feature map, are forwarded through the CNN to be more accurately classified (see text).}\label{fig:PD-proposal}
\end{figure*}

\section{Vision-Based People Detection through Deep Learning}
\label{sec:localization}

For PD, this paper follows the strategy mentioned in the previous section, providing the following contributions. First, we adopt a deep learning based approach, by using pre-trained models.
Second, we are able to drastically decrease the computational effort associated with the exhaustive search performed during the sliding window process.
To accomplish this, we cascade the ACF (non-deep) detector \cite{DollarPAMI2014} with a CNN.
The ACF detector is chosen because it exhibits fast running time figures and high  accuracy when compared to other non-deep detectors, as displayed in  Fig. 10 of the paper \cite{DollarPAMI2014}. According to this work, ACF achieves the runtime of 31.9 FPS and the performance of 17$\%$ log average miss rate in the INRIA dataset, which is the one used in our benchmark experiments.

This proposed cascade strategy is twofold: first, it provides a selective search approach that significantly improves the computational efficiency, since only the output proposals of the ACF are taken into account; then, by cascading with the CNN, we are able to boost the performance of the ACF detector (i.e., improve the classification accuracy of the ACF proposals by reducing the number of false positives).
Fig.~\ref{fig:PD-proposal} illustrates the proposed approach for the PD task.

\subsection{Methodology for PD}
\label{sec:Methodology-for-PD}

In this section we formalize the adopted methodology for PD. First, let us consider that we have available
the following training set ${\cal D}=\{ ({\bf x},{\cal G}_i\}_{i=1}^{|\cal D|}$, where ${\bf x}$ denotes the input image (entire frame) with ${\bf x}:\Omega\rightarrow\mathbb{R}^3$ and $\Omega$ denotes the image lattice\footnote{In this paper, the RGB feature map is considered for the image ${\bf x}$.} of size $w \times h\times d$, with $d=3$; the set of bounding boxes ground truth annotations are defined in ${\cal G}=\{g_k\}_{k=1}^{|{\cal G}|}$, with $g_k=[x^g_k,y^g_k,w^g_k,h^g_k]\in\mathbb{R}^{4}$ denoting the top-left point and width and height (respectively), enclosing the pedestrians.

The training dataset ${\cal D}$ is the input for the \texttt{ACF} detector.
For each input image ${\bf x}_i$, a detector (e.g., \texttt{ACF} or \texttt{LDCF}) is used to provide the candidate windows, or the proposals, along with the scores (confidences). This can be formalized as the following output set ${\cal O}=\{ ({\bf x}({\cal B}), {\cal S})_i\}_{i=1}^{|\cal O|}$. In this set, ${\cal B} = \{b_k\}_{k=1}^{|{\cal B}|}$
represents the set of the detected bounding boxes coordinates, with $b_k=[x^b_k,y^b_k,w^b_k,h^b_k]\in\mathbb{R}^{4}$ denoting the top-left point and width and height enclosing (or not) the pedestrians; we denote as ${\bf x}({\cal B})$, the content (i.e., the proposals, corresponding to a cropped-image) of the image delimited by the bounding boxes ${\cal B}$; ${\cal S}=\{s_k\}_{k=1}^{|{\cal S}|}$ are the detector confidence scores assigned to the proposals ${\bf x}({\cal B})$.

Afterwards, the proposals ${\bf x}({\cal B})$ are further classified resorting to a CNN (details about the CNN formalization are described in \ref{sec:CNN-model}).

As mentioned in \cite{YosinskiNIPS2014}, the generalization ability of the CNN can be boosted resorting to pre-trained models, instead of using random initialization.
Therefore, we use the proposed VGG (Very Deep 16) CNN model
\cite{SimonyanICLR2015}\footnote{Details are also available at: \url{http://www.robots.ox.ac.uk/~vgg/research/very_deep/}.}, pre-trained with Imagenet \cite{RussakovskyIJCV2015}
\footnote{Since we are able to achieve the intended results with the adopted model, both in terms of runtime and accuracy, our system is adequate for HAN tasks as presented in Sec. \ref{sec:used-constraints}. Thus, we did not conduct extensive and exhaustive experiments regarding the CNN architecture, because our focus is directed towards the suitability of the integration of our PD module in the HAN application.}. Formally, we have a dataset to pre-train the CNN, i.e., $\widetilde{\cal D}=\{ (\widetilde{\bf x},\widetilde{\bf y})_n \}_{n=1}^{|{\widetilde{\cal D}}|}$, with $\widetilde{\bf x}:\Omega\rightarrow\mathbb{R}^3$ and $\widetilde{\bf y}\in\widetilde{\cal Y}=\{0,1\}^{\widetilde C}$, where $\widetilde{\bf x}$ are the images, $\widetilde{\bf y}$ are the labels, and $\widetilde C$ is the number of classes in the pre-trained model (in the Imagenet case, the number of classes is $\widetilde{C}=1000$).

In order to re-train (fine-tune) the pre-trained CNN model to the PD task, another dataset is collected based on proposals (cropped-images), i.e., ${\cal D}_{CNN} =\{ ({\bf x}({\cal B}),{\bf y}\}_{i=1}^{|{\cal D}_{CNN}|}$, outputted by a detector (e.g. \texttt{ACF}), and where ${\bf y}\in{\cal Y}=\{0,1\}^{C \times |{\cal B}|}$ denotes the (absence) presence of the pedestrian in the proposals (cropped-images) ${\bf x}({\cal B})$ (i.e., $C=2$). In Section \ref{sec:Adaptation-of-the-deep-network}, we detail how transfer learning and fine-tuning are accomplished for our classification problem.

\section{Materials and methods}\label{sec:Material-and-methods}

This section addresses the implementation details of the adopted deep
learning methodology, to be integrated in the navigation
setup. First, in Sec. \ref{sec:deep-trn-tst} we describe the
experimental setup used to train the deep CNN. Then,
Sec.
\ref{sec:Adaptation of the CNN} addresses the adaptation of the
pre-trained model to the PD task. The outcome of these sections,
i.e., the final CNN architecture fine-tuned for PD, is subsequently
used for testing purposes, as described in Sec.
\ref{sec:Experimental-results}.

\subsection{CNN training in the INRIA dataset}\label{sec:deep-trn-tst}

To train the CNN model we use the INRIA dataset, which is a common benchmark used for research work in the detection of pedestrians in images\footnote{More details can be found at: \url{http://pascal.inrialpes.fr/data/human/}.}. This dataset comprises 1832 training images, from which 1218 are negative images (i.e., not containing pedestrians) and 614 are positive images (i.e., containing pedestrians). There are 288 test images. In our experimental setup, we have to build not only the training set but also the validation set to fine-tune the pre-trained CNN model. To obtain the positive set, we use the ground truth positive training bounding boxes (i.e., proposals corresponding to the image content delimited by them) ${\cal B}_{\rm pos}=1237$. Data augmentation is then performed for the positive samples using the following two steps:
\begin{enumerate}
	\item{Horizontal flipping over the set ${\cal B}_{\rm pos}$, resulting in a new set, ${\cal B}^{(1)}_{\rm pos}=2474$ (including also
		${\cal B}_{\rm pos}$); and}
	\item{Random deformations (including translation and scale), in the range $R=[0, 5]$ pixels (for the beginning and end) applied on the previous set, ${\cal B}^{(1)}_{\rm pos}$. This allows us to obtain a new set, ${\cal B}^{(2)}_{\rm pos}=4948$.}
\end{enumerate}

To build the negative set ${\cal B}_{\rm neg}$, we extract negative
windows (i.e., proposals) from the negative images using the
strategy mentioned in \cite{RibeiroPR2016} (i.e., employing a
non-fully trained version of LDCF). As a result, a set of 12552
negative windows is obtained, by defining an upper-bound of 18
negative proposals per image. This strategy allows to acquire a
total of 17500 proposals, from which 15751 are used for training
(90$\%$) and 1749 for validation (10$\%$).

\subsection{Adaptation of the pre-trained CNN model}\label{sec:Adaptation of the CNN}
As mentioned in Sec. \ref{sec:Methodology-for-PD}, we use a pre-trained CNN model. As such, some modifications/adaptations are necessary. In the following, we describe the required steps, by first detailing the model used and, then, describing the modifications needed.

\subsubsection{Pre-trained model used:}
For the pre-trained network, we select the VGG Very Deep 16 architecture (VGG-VD16), more specifically, the  ``D'' configuration in \cite{SimonyanICLR2015}\footnote{See additional details at: \url{http://www.robots.ox.ac.uk/~vgg/research/very_deep/}.}.

This CNN architecture receives $224\times 224\times 3$ input images. The network comprises 13 convolutional layers, three fully connected layers and a multinomial logistic regression layer (see Sec. \ref{sec:CNN-model}). There are five max-pooling operations (i.e., non-linear subsampling), which operate on a 2x2 region, reducing the input by a factor of two. These operations appear after layers: 2, 4, 7, 10 and 13. The Rectified Linear Unit (ReLU) is selected to be the non-linearity, applied to the convolutional and fully connected layers. Furthermore, the size of the receptive field is the same for all convolutional layers (i.e., 3x3). However, the number of filters is different as described below:
\begin{itemize}
	\item{There are   64 filters in each of the layers 1 and 2;}
	\item{There are  128 filters in each of the layers 3 and 4;}
	\item{There are  256 filters in each of the layers 5, 6 and 7;}
	\item{There are  512 filters in each of the layers from 8 to 13;}
	\item{There are 4096 filters in each of the layers 14 and 15;}
	\item{There are 1000 filters in layer 16. Each of these filters is associated with a different ILSVRC \cite{RussakovskyIJCV2015} class. After this layer, there is a soft-max function.}
\end{itemize}
The pre-training of the mentioned model was performed with Imagenet \cite{RussakovskyIJCV2015}, having 1000 classes, 1.2 million training, 50 thousand validation and 100 thousand test images.

\subsubsection{Adaptation of the deep network:}\label{sec:Adaptation-of-the-deep-network}
 
The VGG-VD16 model that was previously trained (pre-trained) with Imagenet, is going to be re-trained (fine-tuned) with the INRIA PD dataset. However, before re-training this CNN, it should be adapted to our PD task. First, the PD task requires the use of only two classes (i.e., $C=2$) in order to represent the presence or absence of a pedestrian (instead of the previously considered 1000 classes from Imagenet). Therefore, the layer 16 and the soft-max must be adjusted accordingly.

Besides that, we intend to improve the PD detection speed, while achieving a suitable accuracy. To reach this goal we choose to reduce the expected input size for the CNN from $224 \times 224 \times 3$ to $64 \times 64 \times 3$. In the downscaled and PD adapted version of the network, the CNN feedforward time is $0.0485$ seconds, which is substantially faster than the one from the original network, which is $0.3619$ seconds.

As a result of this action, inference is not possible after the first fully connected layer. In order to solve the aforementioned issues, the dimensions of the parameters of the three fully connected layers should be adjusted according to the new input size ($64 \times 64 \times 3$) and the new number of classes (2). We accomplish this by defining compatible parameters dimensions and randomly initializing them. More specifically, they are obtained from a Gaussian distribution, with zero mean and variance $\sigma^2 =0.01$. The changed network is fine-tuned using the previously mentioned (in Sec. \ref{sec:deep-trn-tst}) positive and negative sets, obtained from the INRIA pedestrian dataset. The process of pre-training the network acts as a regularization procedure, similar to data augmentation.

Regarding the fine-tuning hyperparameters, we use $\varepsilon=10$
epochs, a minibatch of $\xi=100$ samples, a learning rate of $\beta=
0.001$, and a momentum of $\mu=0.9$. No special effort was dedicated
to fine-tune the mentioned hyperparameters.

During test, we run the ACF detector in the 288 INRIA test images in
order to obtain the proposals (i.e., regions potentially containing
pedestrians). Then, we run the proposals through the CNN in order to
classify them as pedestrians or non-pedestrians.

All the experiments will be detailed next (see Sec.
\ref{sec:Experimental-results} and Tabs. \ref{tab:Analysis_FP} and \ref{tab:running-times-2}) and were
obtained with Matlab, running on CPU mode on 2.50 GHz Intel Core
i7-4710 HQ with 12 GB of RAM and 64 bit architecture. The Piotr's
Computer Vision Matlab Toolbox \cite{DollarToolbox} (2014, version
3.40) was employed to execute the ACF method, and to perform the
performance evaluation. Regarding the implementation of the CNN
framework, the MatConvNet toolbox \cite{vedaldi15matconvnet} was
utilized.

\section{Evaluation of the People Detection}\label{sec:Experimental-results}
This section provides the testing results for the evaluation of the proposed PD method. The section is divided into two parts:
\begin{itemize}
	\item{In Sec. \ref{sec:Results-CNN-INRIA}, we describe the performance evaluation, concerning both accuracy and runtime figures of the PD method, when the INRIA dataset is used; and}
	\item{In Sec. \ref{sec:Results-CNN-Corridor-MBOT}, we evaluate the performance of the PD method in real scenarios comprising two sequences (termed herein as ``corridor'' and ``MBOT''). Special attention is given to the details of how we achieve real-time requirements in the detection.}
\end{itemize}

\subsection{Performance evaluation on the INRIA dataset}\label{sec:Results-CNN-INRIA}

To test the accuracy of the PD module, to be integrated in the navigation setup, we first assess its performance in the INRIA dataset. The adopted evaluation metric is the log average miss rate (MR), as proposed in \cite{DollarPAMI2012}. This metric is obtained from nine values in the False Positives Per Image (FPPI) interval $[10^{-2},10^0]$. Since it is a miss rate, lower MR values correspond to superior performances (i.e., improved accuracies).

To demonstrate the effectiveness of the proposed cascade of a non-deep detector (ACF) with the CNN architecture, we
also present in Tab. \ref{tab:Analysis_FP} (field ``INRIA Baseline") the results of the ACF alone and the results of the cascade, in order to notice the improvement achieved.
From the experiments conducted, we achieve a log average miss rate of $16.83\%$ for the ACF detector alone. Cascading the ACF with the CNN we are able to reach $15.13\%$.\footnote{The approach proposed herein, as shown in Fig. \ref{fig:PD-proposal}, is general and can be applied to any other non-deep detector (e.g. Regionlets \cite{Wang2013}, LDCF \cite{NamNIPS2014} or Spatial Pooling \cite{Paisitkriangkrai2014} detectors).}

Tab. \ref{tab:Analysis_FP} shows the number of true positives (TP), false positives (FP) and false negatives (FN), before and after the use of the CNN. Notice that, the number of FP is significantly reduced, while maintaining most of the TP. This indicates that the CNN is successfully discarding FP, allowing to reach performance improvements (and justifying the observed gain).

\begin{table}\caption{Logarithmic average miss rate (MR), frames per second (FPS), number of true positives (TP), false positives (FP), and  false negatives (FN) on the INRIA dataset, for the ACF detector and the ACF+CNN.}\label{tab:Analysis_FP}
	\begin{center}
		\begin{tabular}{|c|c|c|c|}
			\hline
			{\bf  Dataset}  &  {\bf  Metrics}  &  {\bf ACF Proposals}  &  {\bf ACF+CNN}
			\\ \hline
			
			&       \texttt{TP}    &   551   &  546      \\
			INRIA &  \texttt{FP}   &  1284   &  249      \\
			Baseline &\texttt{FN}  &   38    &  43      \\
			&        \texttt{MR\%} &  16.83	&  15.13	\\
			&        \texttt{FPS}  & 15.18	& 3.38  	\\			
			
			\hline
			&       \texttt{TP}    &   531   &  528      \\
			INRIA &  \texttt{FP}   &   112   &  54      \\
			Threshold &\texttt{FN} &   58    &  61      \\
			& \texttt{MR\%}  &  17.48	&  16.48	\\			
			&        \texttt{FPS} & 14.81	& 6.62 	\\
			
			\hline
		\end{tabular}
	\end{center}
\end{table}

An important concern when building the PD algorithm, is to verify and ensure that the runtime figures satisfy the real-time requirements. Tab. \ref{tab:Analysis_FP}
shows the ACF and ACF+CNN running times in frames per second (FPS) obtained for the INRIA dataset.
From the results, we conclude that the ACF+CNN system is able to perform PD at 3.38 FPS, which should be improved. The introduction of the CNN in the pipeline after the ACF method, resulted in a significant runtime decrease from 15.18 FPS to 3.38 FPS.

To improve the mentioned runtime figures, while achieving as much accuracy as possible, two strategies can be used: $(a)$ reduce the original images size, or $(b)$ discard ACF proposals with confidence score below a certain threshold.

In this work, the latter option $(b)$ is adopted. Notice that, reducing the size of the images may jeopardize the quality of the detections. Furthermore, the confidence scores outputted by the ACF detector constitute a relevant indicator to filter the proposals. This can be achieved by applying a threshold over the proposals, where only the ones above a certain score are kept and are further processed by the CNN. For our experiments, a threshold value of 40 was set for the confidence score. This threshold is in accordance with the value suggested in \cite{VermaWICCV2015}, in which it is proposed a thresholding technique based on upper and lower bounds on the confidence scores of the ACF. In \cite{VermaWICCV2015}, the lower threshold value, below which the ACF proposals are discarded, is 38 and was attained resorting to grid search with cross validation. The objective was to find the lowest miss rate, when the false positives per image are equal to 0.1.

Another important remark is that, the gain in speed results from the fact that the CNN only has to classify a smaller portion of the ACF proposals, instead of all of them. Therefore, the easier false positives should be discarded by the threshold operation, while the harder false positives should be discarded by the CNN. The threshold value controls the trade-off between potential accuracy loss and speed gain.

By selecting the later case (i.e., the threshold operation), we are able to improve the runtime of the overall detector (i.e., ACF+CNN), when compared with the results obtained for the baseline. These results are shown in Tab.~\ref{tab:Analysis_FP} (field ``INRIA Threshold"), where it can be seen that the ACF+CNN frame rate increases from 3.38 FPS (baseline) to 6.62 FPS (threshold).

In order to assess if this speed metrics are suited for real-time applications, we perform further experiments in real HAN scenarios in Sec. \ref{sec:Results-CNN-Corridor-MBOT}.

\subsection{Performance evaluation on real scenarios}\label{sec:Results-CNN-Corridor-MBOT}

\begin{figure*}
	\vspace{-0.25cm}
	\centering
	\subfloat[Pedestrian detection in the ``corridor'' sequence.]
	{\includegraphics[width=0.32\textwidth]{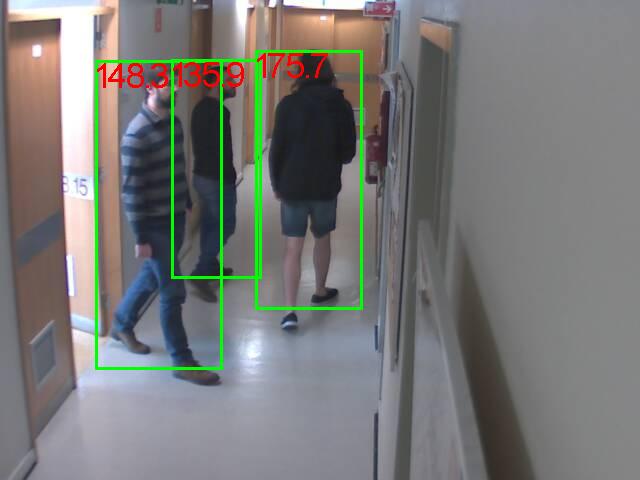}
		\includegraphics[width=0.32\textwidth]{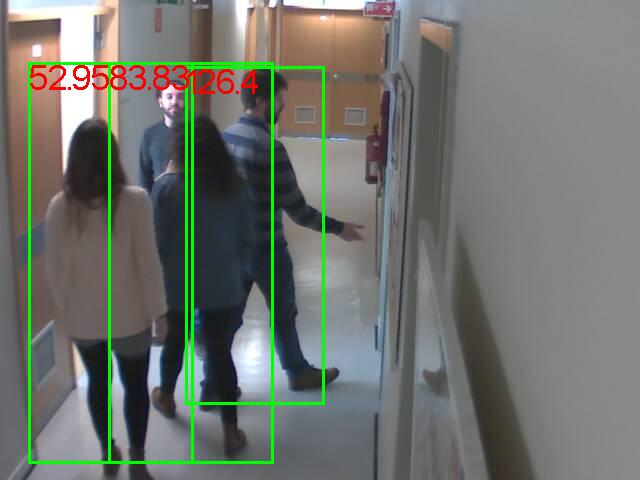}
		\includegraphics[width=0.32\textwidth]{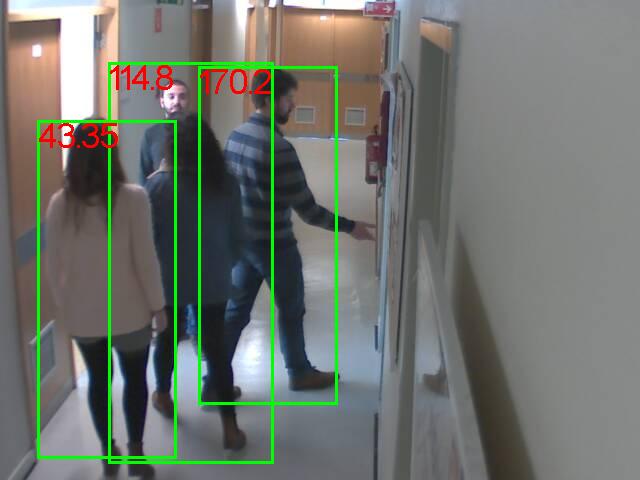} \label{fig:david1}}\\
	\subfloat[Pedestrian detection in the ``MBOT'' sequence.]
	{\includegraphics[width=0.32\textwidth]{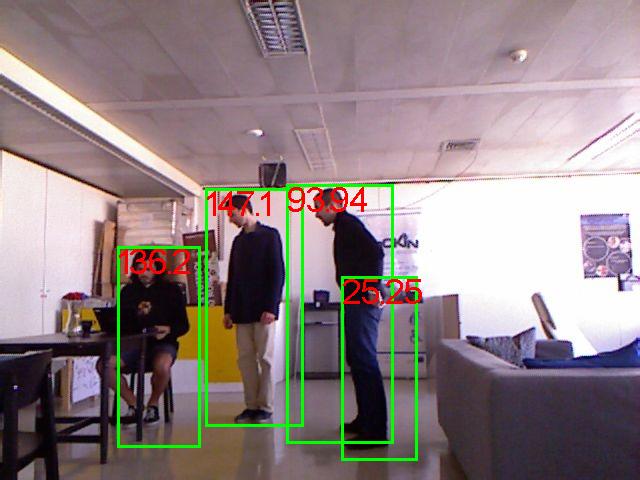}
		\includegraphics[width=0.32\textwidth]{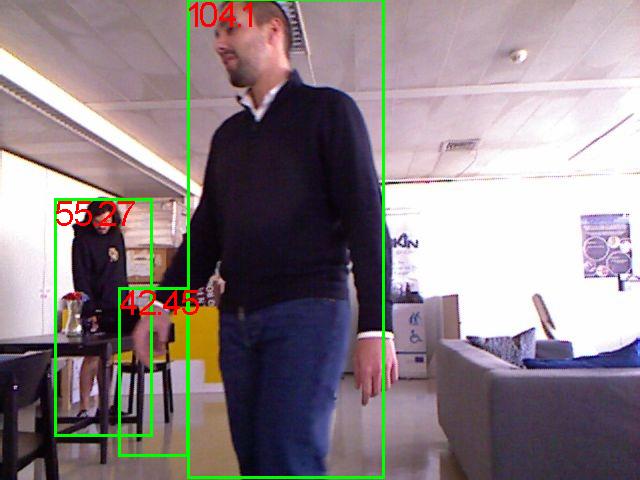}
		\includegraphics[width=0.32\textwidth]{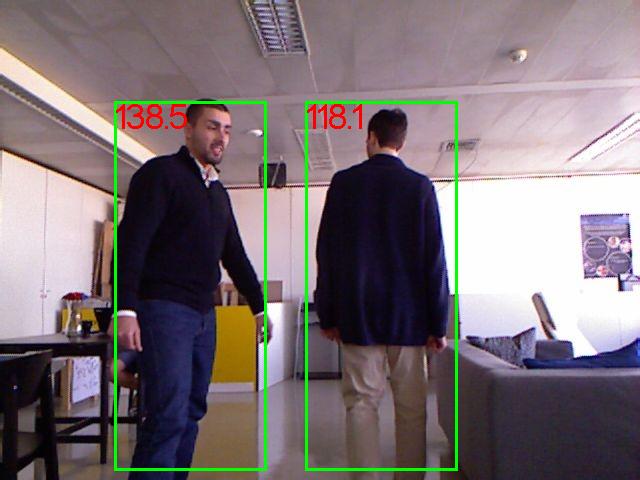}\label{fig:david2}}\\
	\caption{In real scenarios, there are cameras mounted on the ceiling and on the robot. To test our PD, we used two sequences of images acquired from both possible real scenario camera locations. In Figs.~\protect\subref{fig:david1} and~\protect\subref{fig:david2} are shown three images of the ``corridor'' and ``MBOT'' sequences, respectively.}
	\label{fig:david-results}
\end{figure*}

To perform experiments in real scenarios, we acquired two indoor datasets to evaluate the PD task.
Two datasets are considered in these experiments:
\begin{itemize}
	\item{The ``corridor'' dataset, containing 5556 images; and}
	\item{The ``MBOT'' dataset, containing 3966 images.}
\end{itemize}
The size of the frame (i.e., the image) is $480\times 640$ for both of the sequences. Some of the results obtained in the two datasets are shown in Fig. \ref{fig:david-results}, where each bounding box has a score, showing the confidence of containing a pedestrian.

The total time to perform PD in the dataset ``corridor'' (5556 frames) and ``MBOT'' (3966 frames), is  roughly  707.27 seconds and 839 seconds, respectively. The running time figures per frame are shown in Tab. \ref{tab:running-times-2}. In this table, the field ``Baseline'' refers to the metrics of the final PD detector as detailed in Sec. \ref{sec:Adaptation of the CNN}. The algorithm reaches 7.85 FPS and 4.84 FPS, for the ``corridor'' and the ``MBOT'' sequences, respectively.

The runtime is longer for the ``MBOT'' dataset, because the number of ACF detections that have to be processed by the CNN is larger, in comparison with the ``corridor'' dataset. This observation is depicted in Tab.~\ref{tab:running-times-2}, by comparing the average number of detections (``avg. no. det.'') before (ACF only) and after (ACF+CNN) the application of the CNN, and also the CNN time for each data sequence (top, the two columns in the field named ``Baseline'').

By applying the threshold operation (as described in Sec. \ref{sec:Results-CNN-INRIA}), it possible to obtain approximately 10 FPS for each of the datasets (as shown in Tab. \ref{tab:running-times-2}, field   ``Threshold'')), which is suited for real-time applications.

To assess the advantages of using the cascade ACF+CNN (including the threshold operation), instead of exhaustive search CNN classification (for single scale detection), we conduct an analysis to determine the speed that could be achieved by the latter. The details of this study are presented in \ref{sec:comp_cascades_exhaustive_search}. As a result, we reached a runtime of 0.1238 FPS for the exhaustive search, considering only a single scale, which is not suitable for HAN tasks, and is significantly worse than the speed achieved by our proposed method, using multiple scales, which is approximately 10 FPS (as mentioned in Tab.~\ref{tab:running-times-2}, or 6.62 FPS in INRIA, as presented in Tab.~\ref{tab:Analysis_FP}).

\newcommand{\scz}{\scriptsize}

\begin{table}
	\caption{Runtime figures before (top, ``Baseline") and after (bottom, ``Threshold") the threshold operation applied to the ACF proposals, when using the overall PD method (i.e., ACF+CNN).}
	\label{tab:running-times-2}
	\begin{center}
		\begin{tabular}{|c|c|c|}
			\hline
			{\bf  Dataset} & {\bf  Data seq. 1 (corridor)} & {\bf Data seq. 2 (Mbot)}   \\ \hline
			
			&              \scz Total time       =  0.1273 sec. &    \scz Total time  =  0.2066 sec. \\
			& 			 \scz avg. no. det. ACF = 2.70  		 &   \scz avg. no. det. ACF = 4.83  \\
			& 			 \scz avg. no. det. ACF+CNN = 2.36   	     &    \scz avg. no. det. ACF+CNN = 2.45  \\
			Baseline    &  \scz ACF time         =  0.0326 sec.       &    \scz ACF time    =  0.0367 sec.       \\
			&              \scz CNN time         =  0.0947 sec.       &    \scz CNN  time   =  0.17   sec.       \\
			&              \scz {\bf Frame rate} =   {\bf 7.85} FPS   &    \scz {\bf Frame rate} =   {\bf 4.84}  FPS \\
			\hline
			
			&              \scz Total time  =  0.0961 sec.            &    \scz Total time  =  0.1026 sec.       \\
			& 			 \scz avg. no. det. ACF = 1.82   			   &  \scz   avg. no. det. ACF = 1.78 \\
			& 			 \scz avg. no. det. ACF+CNN = 1.81    		   &  \scz   avg. no. det. ACF+CNN = 1.73  \\
			Threshold   &  \scz ACF time    =  0.0333 sec.            &    \scz ACF time    =  0.0381 sec.       \\
			&              \scz CNN time    =  0.0628 sec.            &    \scz CNN  time   =  0.0645 sec.       \\
			&              \scz {\bf Frame rate} =   {\bf 10.41} FPS  &    \scz {\bf Frame rate} =   {\bf 9.74} FPS  \\
			\hline
		\end{tabular}
	\end{center}
\end{table}

\section{Human-Aware Navigation}
\label{sec:used-constraints}
From the method described in the previous section, we get bounding boxes on the images, representing pedestrians. Multiple instances of the proposed PD can be applied in images of several cameras at the same time. Assuming that we have a set of images (a single image can also be used), the goal of this section is to: firstly, project the position of the pedestrians from the image into the world environment; fuse the information given from different imaging sensors; and, then, define the HAN constraints to be included in any conventional path planer.

After computing the position of the pedestrian in the world coordinate systems, including the estimation of the pedestrian's velocity (the methods used in this paper for this module are shown in Sec.~\ref{sec:person_tracking}), the main goal of this section is to define the robot's path in an environment with humans that may interact with it. To be as Human-Aware as possible, three goals are considered, namely: human comfort; respect social rules; and naturalness. To fulfill these requirements, the following constraints are taken into account:
\begin{enumerate}
	\item{Take least effort path (naturalness) -- Sec.~\ref{sec:path_planer};}
	\item{Keep a distance from static obstacles (naturalness) -- Sec.~\ref{sec:path_planer};}
	\item{Respect personal spaces (human comfort) -- Sec.~\ref{sec:personal_space_constraint};}
	\item{Avoid navigating behind sitting humans \\ (human comfort) -- Sec.~\ref{sec:visibility_constraint};}
	\item{Do not interfere with human-object interactions (human comfort) -- Sec.~\ref{sec:iteraction_constraint}; and }
	\item{Overtake people by the left (social rule) -- Sec.~\ref{sec:overtake_constraint};}
\end{enumerate}
The first two constraints are related to navigation problems (the method used in this paper is described in Sec.~\ref{sec:path_planer}), whereas the remaining four constraints are about the HAN (more details about each of these constraints are shown in Sec.~\ref{sec:human_aware_constraints}).

\subsection{People Tracking in the World} \label{sec:person_tracking}
As explained in the previous sections, the PD module returns bounding boxes representing people in the scene. Those are sent to this module which uses the Nearest Neighbour Joint Probabilistic Data Association (NNJPDA), \cite{Bar2009}, and an array of Kalman Filters to track people (one for each person). It performs the following steps:
\begin{enumerate}
	\item{Project the middle point, between the bottom left and right corners of each bounding box, to the ground plane, in world coordinates;}
	\item{Predict the new state of each track;}
	\item{Associate the measurements to existing tracks and determine if tracks need to be created or removed;}
	\item{Update the prediction with the respective measurement; and}
	\item{Create and/or remove signaled tracks.}
\end{enumerate}

The people tracking is performed in the world coordinate system, instead of the image plane. Consider the world coordinate system to be the position in the $x$ and $y$ coordinates in the floor plane ($z=0$). Each measurement is defined as $\mathbf{m} = [m_x,m_y]^T$. For this purpose, two assumptions are considered: i) people are standing or walking upright\footnote{In our experiments we also considered a seated person. Even though our method was not designed for this type of cases, the results were satisfactory.}; ii) given i), a person's feet will always be on the ground plane and, thus, the point, which represents a person feet, is on the line between the bottom left and right corners of the respective bounding box. Since the person is, most of the times, in the center of the bounding box, a good estimate for the position of the feet is in the middle point of the considered line segment. The projection is performed by transforming the selected point with an homography computed a priori (transformation from the image plane into the floor plane). These positions will be associated with the targets and be used as measurements in Kalman Filters (one for each person), which will perform the tracking. A constant velocity motion model was considered for the prediction step of the filter. The state of a person is considered to be the position and the velocity, which we define as $\mathbf{p} = [p_x,p_y,v_x,v_y]^T$. Velocity estimates (required for the HAN constraints) are given from the Kalman filter state estimates, the person orientation is then defined as
\begin{equation}
    \alpha_p = atan2(v_y,v_x).
    \label{eq:person_orientation}
\end{equation}

The association between measurements and targets is performed with the NNJPDA method \cite{Bar2009}. This is a hard assignment method, in which only one measurement is assigned to the target, using a Maximum A Posteriori (MAP) approach. When the association is complete, there is a condition that checks if the assignment distance is higher than a threshold. When that is the case, the track and measurement are unassigned. After the assignment, there are three cases to be considered:
\begin{enumerate}
	\item{The assignment exists;}
	\item{The track is not assigned; and/or}
	\item{The measurement is not assigned.}
\end{enumerate}
If there is a successful assignment, it proceeds to the correction step of the filter, with the assigned measurement. The resulting state is then evaluated to define if the person is standing, walking or seated. A person is considered to be standing if norm of the velocity estimate is smaller than $0.1m/s$. If besides satisfying the previous threshold and it is in one of the defined a priori sitting areas, then the person is considered to be seated. Finally if the norm of the velocity estimate is higher than the threshold the person is assumed to be walking.
When a track is not assigned, there are two possibilities: the track increases the inactivity flag and, if the inactivity threshold was reached, the track is deleted. If a measurement is not assigned, after some iterations, a new Kalman Filter is created for this measurement.

\begin{figure*}
	\vspace{-0.25cm}
	\centering
	\subfloat[Personal space]{\includegraphics[width=0.25\textwidth]{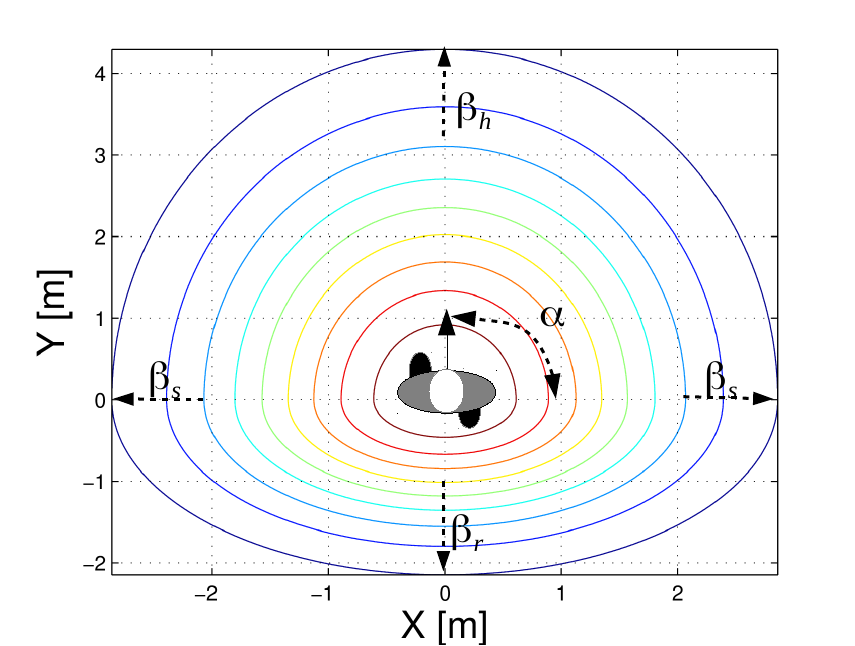}\label{fig:as_gauss:moving}}\hfill
	\subfloat[Object hand over]{\includegraphics[width=0.25\textwidth]{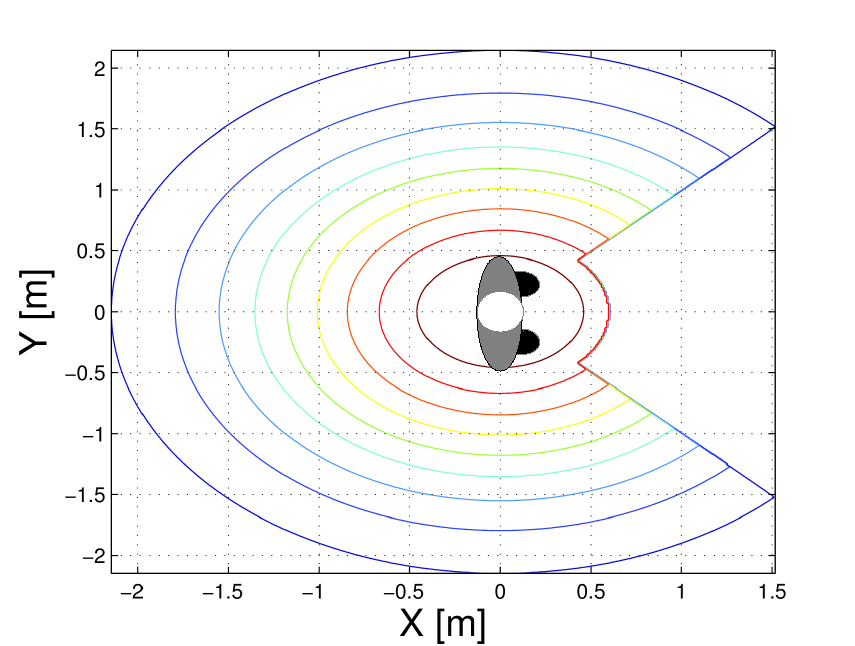}\label{fig:as_gauss:stop}}\hfill
	\subfloat[Seated person]{\includegraphics[width=0.25\textwidth]{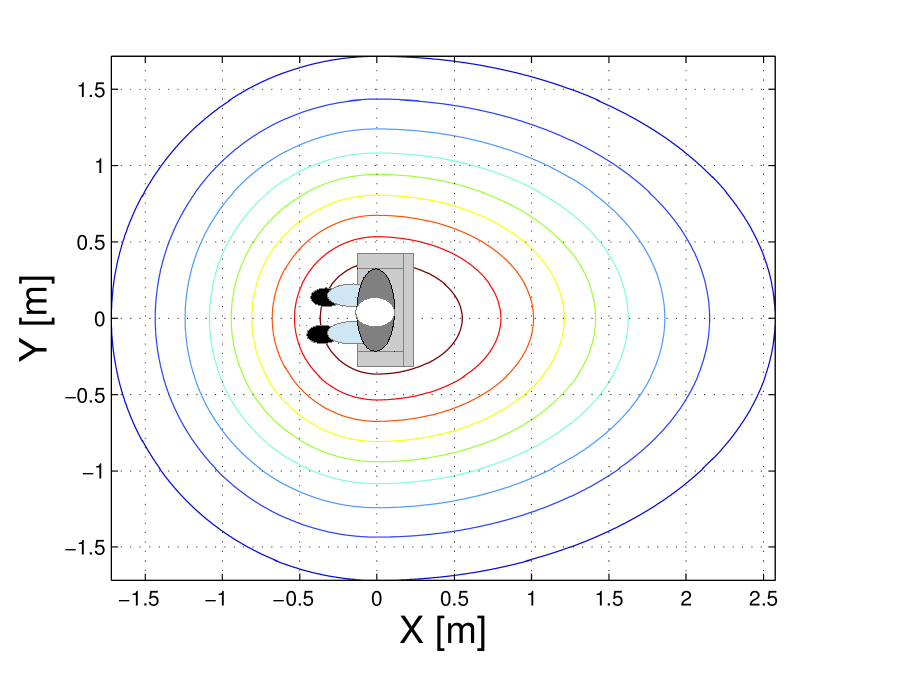}\label{fig:as_gauss:seated}}\hfill
	\subfloat[Walking person]{\includegraphics[width=0.25\textwidth]{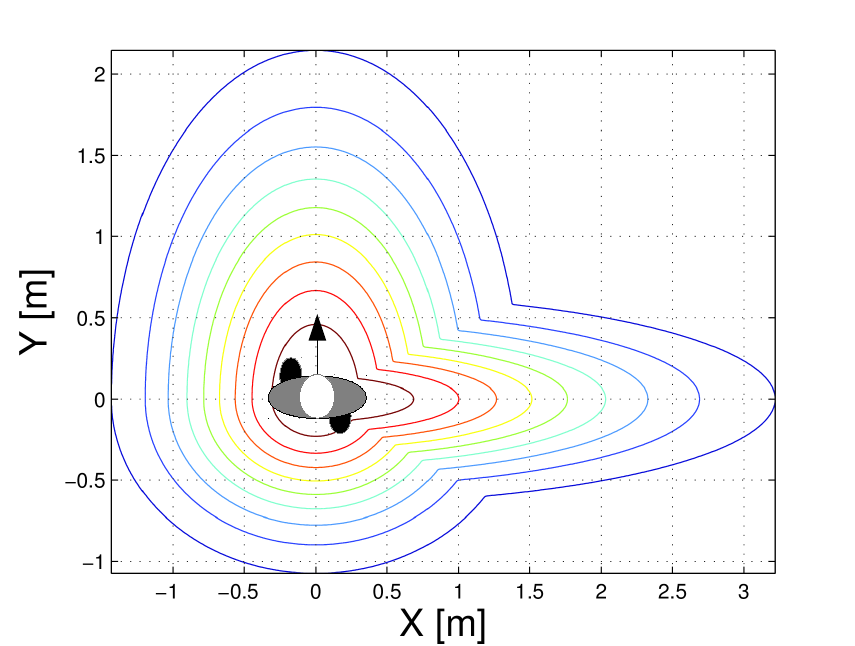}\label{fig:as_gauss:walking_L}}
	\caption{Representation of cost functions associated with different people postures: Fig.~\protect\subref{fig:as_gauss:moving} represents the cost function for the personal space of a person walking in the $y$ direction at 1~[m/s]; Fig.~\protect\subref{fig:as_gauss:stop} shows the cost function of a person standing, oriented in the $x$ direction, during an object hand over; Fig.~\protect\subref{fig:as_gauss:seated} represents the cost function for the case where a person is seated; and Fig.~\protect\subref{fig:as_gauss:walking_L} shows the total cost function of a walking person, including the social rule of overtaking her by the left and her personal space.}
	\label{fig:as_gauss}
\end{figure*}

\subsection{Path planner and obstacle avoidance}\label{sec:path_planer}
The first HAN constraint is addressed by the path planner (see the list in the beginning of Sec.~\ref{sec:used-constraints}). In this work the $A^*$ algorithm \cite{hart:TSSC1968} was used, ensuring a minimum cost path as long as the heuristic is admissible. The total cost of a node is given by the sum of the cost of reaching that node, with the heuristic cost. The latter was considered to be the Euclidean distance from the initial to the goal position. Since the environment is dynamic (people may appear walking in the scene), the planner computes a path periodically.

The goal of the second constraint is to prevent the robot from passing too close to obstacles. This is solved by attributing a high cost to the area surrounding the obstacles, \cite{article:marder:2010}.

In the next subsection, we define the remaining constraints, that will be included in the path planner, for the navigation to be human-aware.

\subsection{Human-Aware Navigation Cost Functions}\label{sec:human_aware_constraints}
The Human-Aware Cost Functions used in this work are based on previous state-of-the-art approaches. However, some of them are reformulated (namely the cost functions associated with constraints 4 and 5) in order to be better integrated in our approach, and to standardize their formulation.

\subsubsection{Personal Space Cost Functions:}\label{sec:personal_space_constraint}
The third constraint in the list accounts for personal space, which models will be presented next. We consider three different situations: when a person is standing; walking; or seated. For the case of a walking person, we used the formulation proposed in \cite{34}, which the authors call asymmetric Gaussian:
\begin{equation}
g = \text{asymGauss}(e_x,e_y,p_x,p_y,\alpha,\beta_h,\beta_s,\beta_r),
\end{equation}
where
\begin{itemize}
	\item{$\alpha$  -- orientation of the function;}
	\item{$\beta_h$ -- variance in the $\alpha$ direction;}
	\item{$\beta_r$ -- variance in the $\alpha - \pi$ direction;}
	\item{$\beta_s$ -- variance in the $\alpha \pm \frac{\pi}{2}$ direction;}
	\item{$e_x$ and $e_y$ -- are the variables that define the space around the person; and}
	\item{$p_x$ and $p_y$} -- represent the person position.
\end{itemize}
A graphical representation of these parameters is shown in Fig.~\ref{fig:as_gauss}\subref{fig:as_gauss:moving}. Then, the personal space of a walking person was modelled as:
\begin{multline}
g_1 = \text{asymGauss}(e_x,e_y,p_x,p_y,\alpha_p,\widetilde{\beta},\frac{2}{3}\widetilde{\beta},\frac{1}{2}\widetilde{\beta}), \\ \text{where} \; \widetilde{\beta} = \text{max}\left(\nu, 0.8\right),
\label{eq:f1}
\end{multline}
$\alpha_p$ is the person's orientation and $\nu$ her speed (norm of the velocity vector). A graphical representation of the person walking along y--axis direction with a velocity of 1 [m/s] is presented in Fig.~\ref{fig:as_gauss}\subref{fig:as_gauss:moving}.

\begin{figure*}
	\vspace{-0.25cm}
	\subfloat[The robot was placed behind a person and a goal position was defined. Then the person starts walking, the robot should replan a path through the person's left.]{\includegraphics[width=0.245\textwidth]{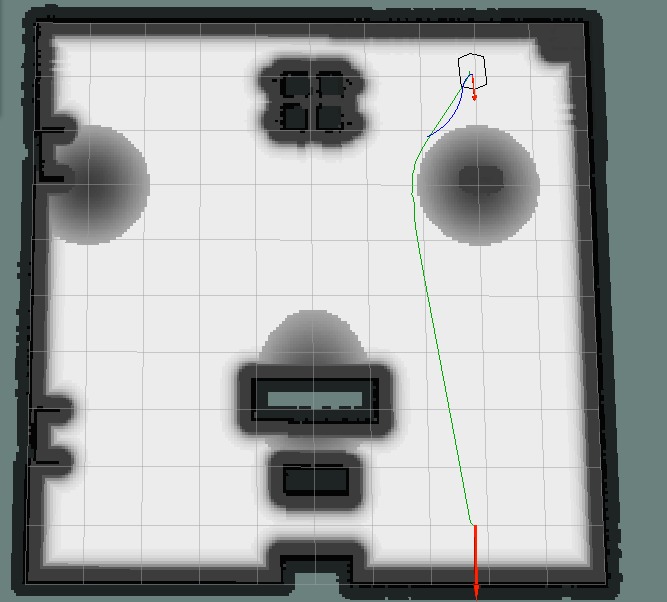}
		\includegraphics[width=0.245\textwidth]{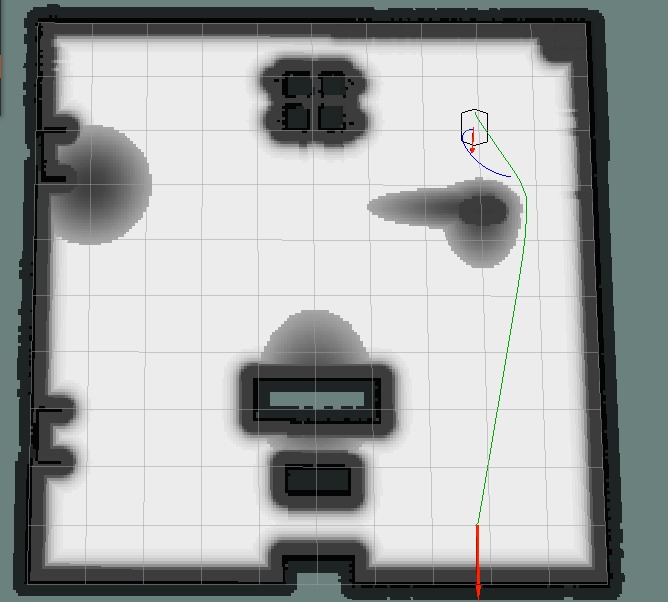}
		\includegraphics[width=0.245\textwidth]{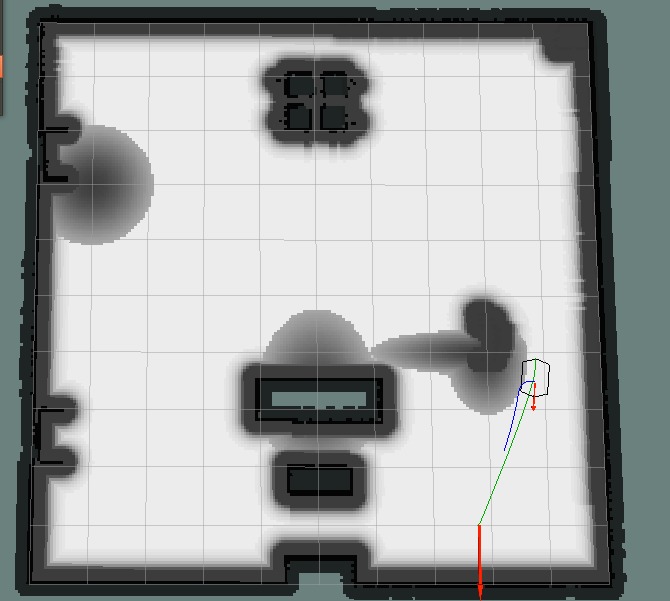}
		\includegraphics[width=0.245\textwidth]{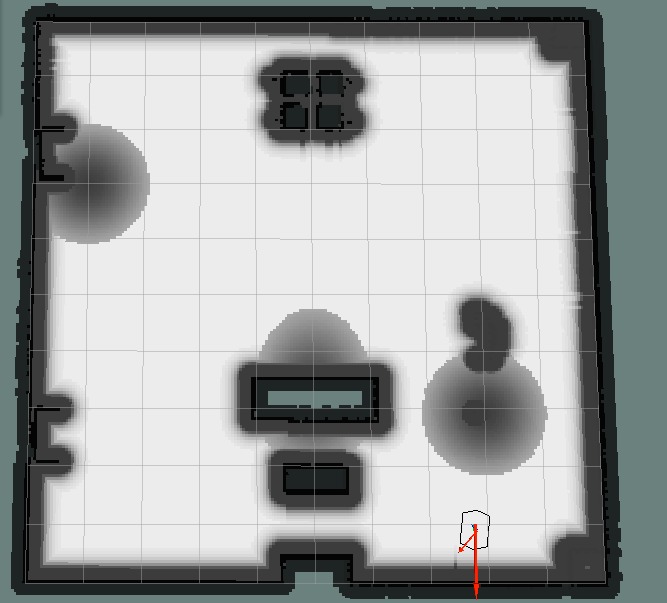}\label{fig:simulated_environment:exp3}}\\
	\subfloat[The robot is requested to hand over an object to a person, who is across the room. However it needs to pass between a seated person and a TV. As it starts moving, the TV is turned on. To prevent interference, the robot replans around that area. However, there is yet another person, walking behind the couch, who must be taken into account.]{\includegraphics[width=0.245\textwidth]{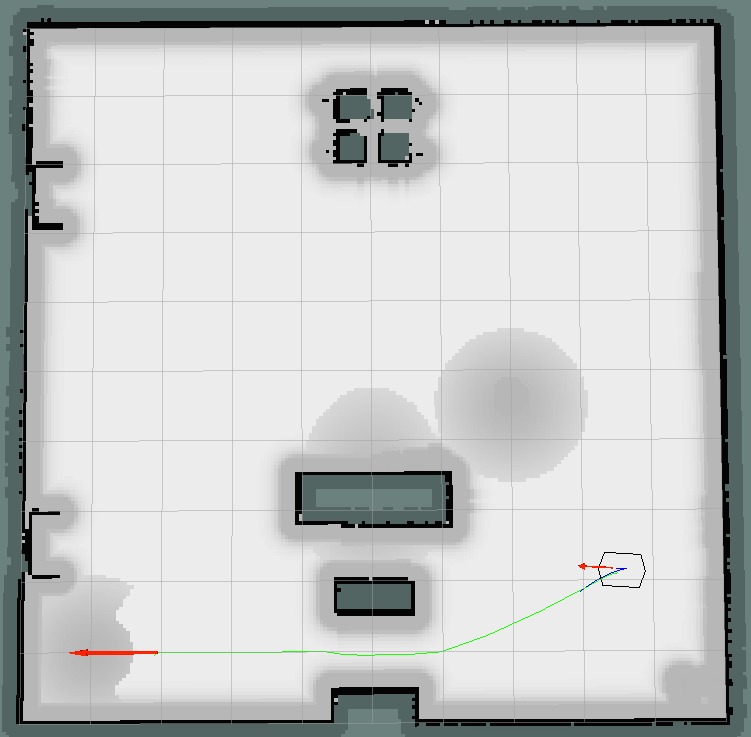}
		\includegraphics[width=0.245\textwidth]{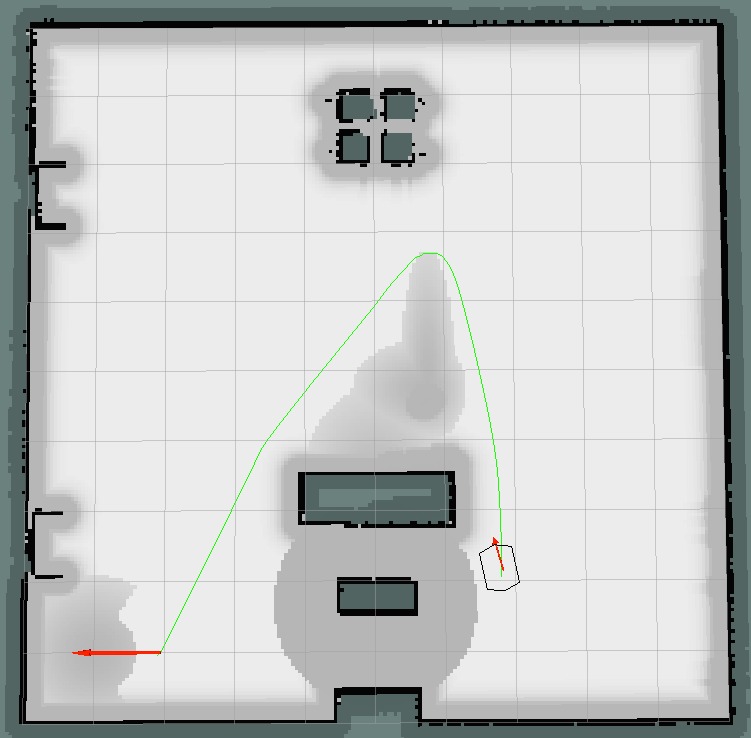}
		\includegraphics[width=0.245\textwidth]{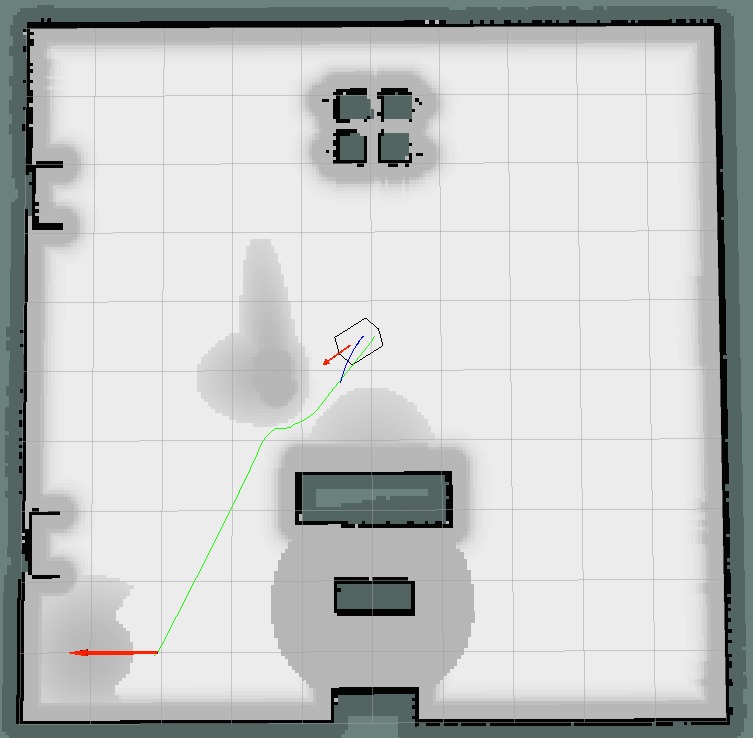}
		\includegraphics[width=0.245\textwidth]{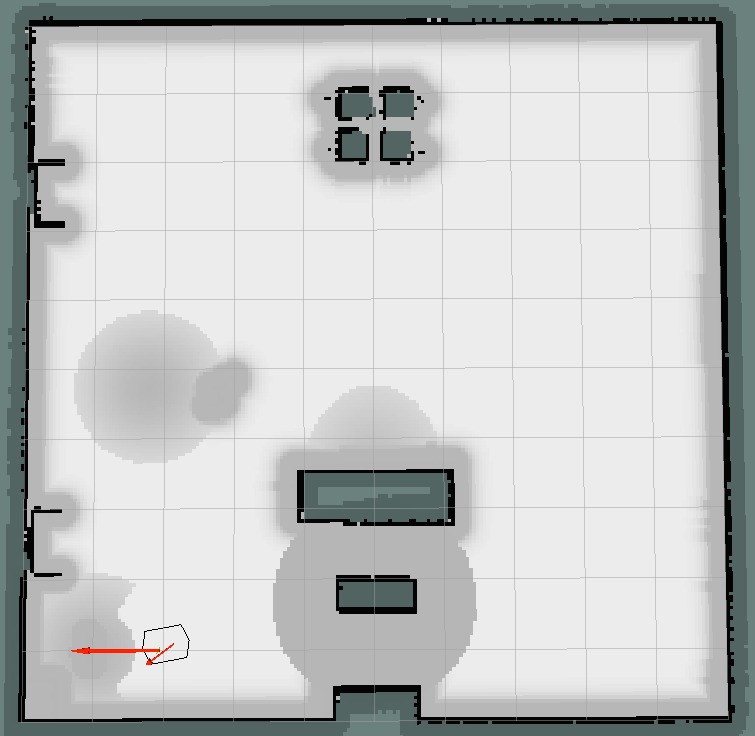}
		\label{fig:simulated_environment:exp5}}\\
	\caption{Evaluation of the proposed navigation system using simulated environments. The environment was created using Gazebo and the results are shown in Rviz (ROS package). Figs.~\protect\subref{fig:simulated_environment:exp3} 
		and~\protect\subref{fig:simulated_environment:exp5} show sequences of images representing experiments 1 and 2 respectively (more details regarding each of the experiments are given in the text). In all cases, it can be seen: the costmap, current and goal position, path, and the trajectory.}
	\label{fig:simulated_environment}
\end{figure*}

Regarding a walking person, it makes sense for the personal space in front to be larger than in the back (to ensure that the robot does not pass in front of the person, decreasing the risk of collision). On the other hand, if a person is standing and we consider the personal space defined using the previous formulation, the robot may pass behind too close to the person, causing discomfort. Thus, for this case we propose that the personal space to be modelled as a circular Gaussian:
\begin{equation}
g_2 = \text{exp}\left(-\frac{\left(e_x-p_x\right)^2}{2\beta_{x}^2} - \frac{\left(e_y -p_y\right)^2}{2\beta_{y}^2}\right),
\label{eq:f2}
\end{equation}
where $\beta_{x}$ and $\beta_{y}$ are the standard deviation in the x and y direction respectively. This formulation was also considered for a seated person.

If an object hand over is required, the robot should be able to enter the personal space, to be at ``arm's length''. However, the robot cannot be allowed to enter from a random direction, instead it should only be allowed to approach a person from the front \cite{article:koay:2007}. Thus, our solution is to open the region in front of the person 45 degrees, to a distance of 0.6[m]. Personal space, in a hand over scenario, is depicted in Figure~\ref{fig:as_gauss}\subref{fig:as_gauss:stop}.

\subsubsection{Visibility Constraint:}\label{sec:visibility_constraint}
Constraint 4 concerns preventing discomfort from passing behind a seated person. We reformulated this problem with an asymmetric Gaussian:
\begin{equation}
g_3 = \text{asymGauss}(e_x,e_y,p_x,p_y,\alpha_p-\pi,1.2,0.8,0.006).
\label{eq:f3}
\end{equation}

\begin{figure*}
	\centering
	\begin{tabular}{cccc}
		\multirow{2}{*}[7.5em]{\subfloat[Image of the robot platform used for the experimental results in realistic scenarios.]{\includegraphics[width=0.28\textwidth]{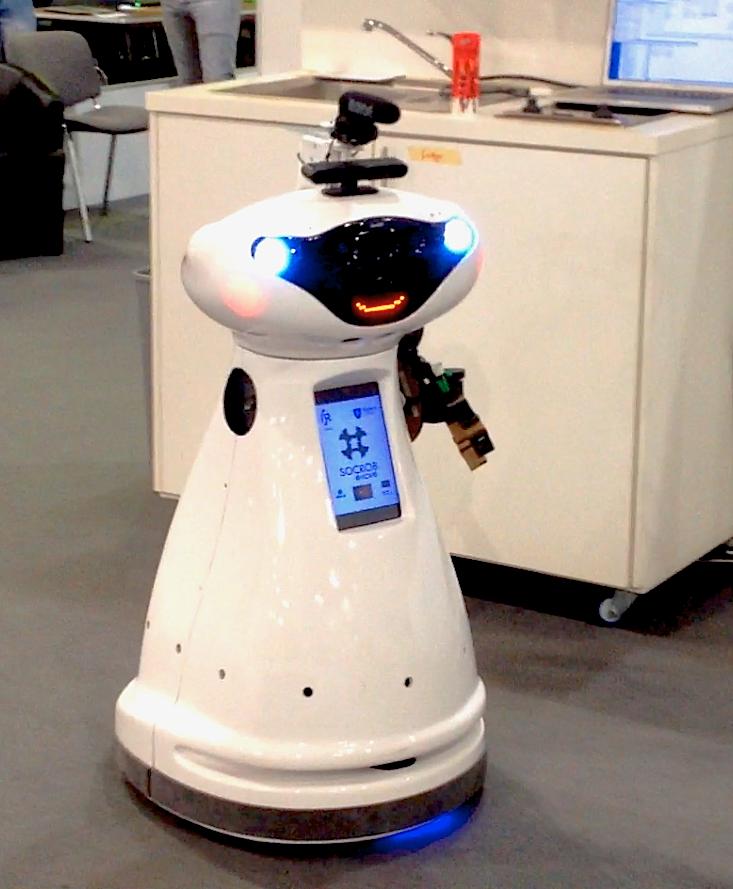}\label{fig:real_setup:mbot}}} &
		\subfloat[Image of {\tt Camera 1}, mounted on the ceiling.]{\includegraphics[width=0.19\textwidth]{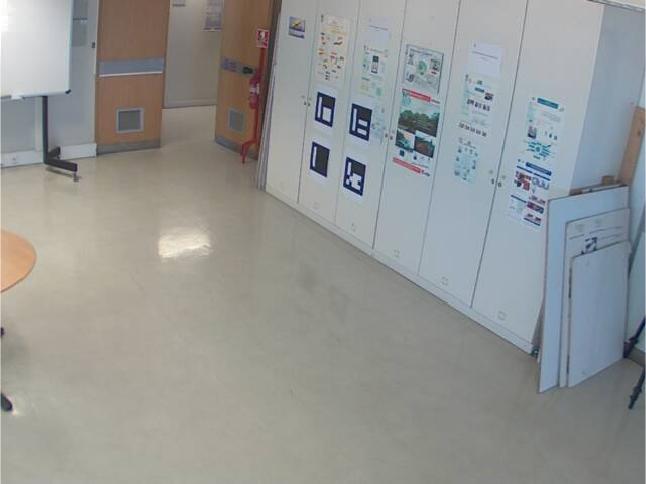}\label{fig:real_setup:camera1}} &
		\subfloat[Image of {\tt Camera 2}, mounted on the ceiling.]{\includegraphics[width=0.19\textwidth]{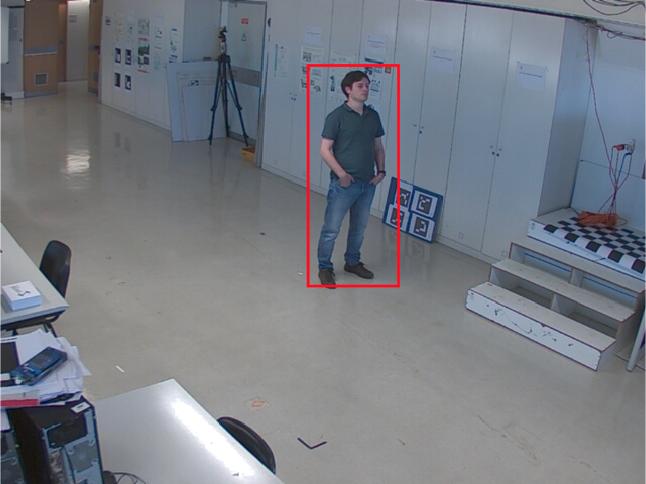}\label{fig:real_setup:camera2}} & \\ &
		\subfloat[Image of {\tt Camera 3}, mounted on the ceiling.]{\includegraphics[width=0.19\textwidth]{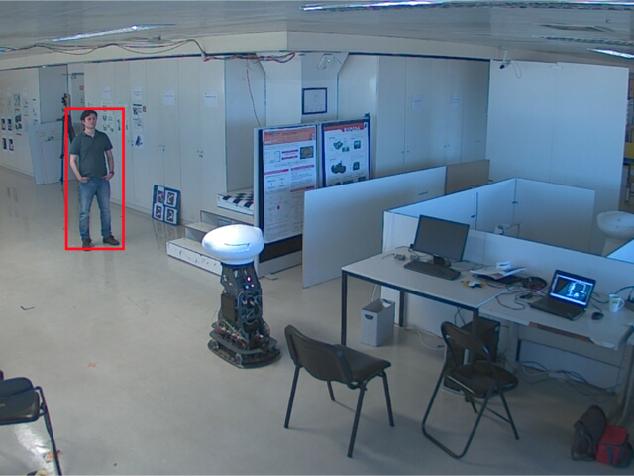}\label{fig:real_setup:camera3}} &
		\subfloat[On-board robot camera's image.]{\includegraphics[width=0.19\textwidth]{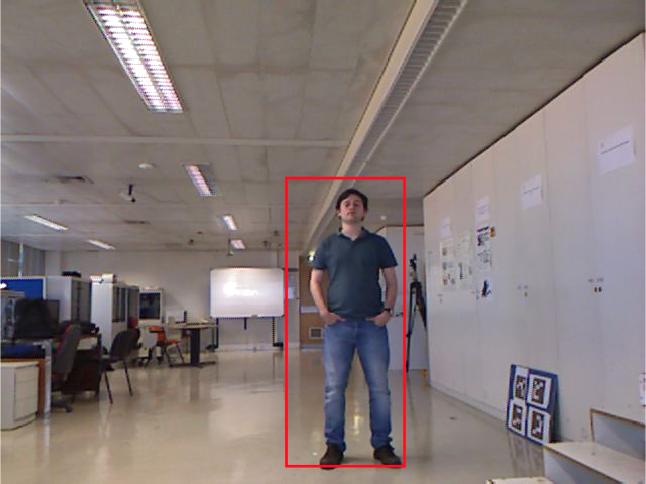}\label{fig:real_setup:camera4}} &
		\multirow{-2}{*}[16.8em]{\subfloat[Depiction of the environment in Rviz, showing: the robot's position, a pedestrian, and the positions of the cameras.]{\includegraphics[width=0.245\textwidth]{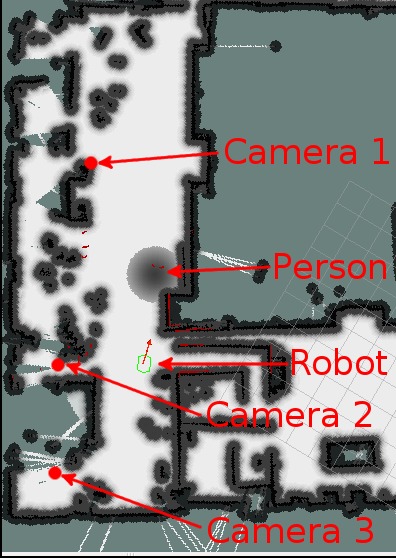}\label{fig:real_setup:rviz}}}
	\end{tabular}
	\vspace{0.25cm}
	\caption{Representation of the setup used in the experiments in a realistic scenario. Fig.~\protect\subref{fig:real_setup:mbot} shows the robot platform and Figs.~\protect\subref{fig:real_setup:camera1},~\protect\subref{fig:real_setup:camera2},~\protect\subref{fig:real_setup:camera3}, and~\protect\subref{fig:real_setup:camera4} show the images of the cameras that will be used to detect the pedestrians (as it can be seen, in these images we already show the bounding boxes identifying a person in the environment). To conclude, Fig.~\protect\subref{fig:real_setup:rviz} shows the environment (ROS Rviz package), with the position of all the cameras, the position of the robot and the pedestrian, with the respective HAN constraint (in this case the pedestrian was standing).}
	\label{fig:real_setup}
\end{figure*}

\subsubsection{Interaction Constraint:}\label{sec:iteraction_constraint}
The fifth constraint prevents the robot from interfering with a person interacting with an object. It is represented by an interaction set modelled as a circle:
\begin{equation}
g_4 = \left\{ \begin{array}{l l} \gamma & \quad \text{if} \quad \left(e_x-\widetilde{p}_x\right)^2 + \left(e_y-\widetilde{p}_y\right)^2 \leq r\\ 0 & \quad \text{otherwise} \end{array} \right. ,
\label{eq:f4}
\end{equation}
where the middle position between the interacting entities is denoted as $\left(\widetilde{p}_x,\widetilde{p}_y\right)$, and the radius $r$ is half the distance between the entities (only one-on-one interactions are considered). $\gamma$ is an importance factor, which varies from 0 to 1.

\subsubsection{Overtake Constraint:}\label{sec:overtake_constraint}
Constraint 6 represents the social rule of overtaking people by the left (considered only for walking persons). This constraint is also represented using an asymmetric Gaussian:
\begin{equation}
g_5 = \text{asymGauss}(e_x,e_y,p_x,p_y,\alpha_p-\frac{\pi}{2},1.5,0.3,0.0075),
\label{eq:f5}
\end{equation}

\subsubsection{Cost Function Fusion:}
For the three possible postures of a person (standing, seated and walking), there are two where more than one cost function is applied and they must be combined. Since the main goal of the framework is to maximize the comfort of the humans, the cost functions are combined by taking the maximum cost value attributed to each point in space. The first case of multiple cost functions affecting the same space, is a seated person, whose personal space is given by:
\begin{equation}
g_6 = \max\left(g_2, g_3\right),
\end{equation}
this cost function is depicted in Fig.~\ref{fig:as_gauss}\subref{fig:as_gauss:seated}. The second case concerns a walking person, where the personal space must be combined with the respective social rule (a person should be overtaken by the left):
\begin{equation}
g_7 = \max\left(g_1, g_5\right).
\end{equation}
A graphical representation of this cost function is shown in Fig.~\ref{fig:as_gauss}\subref{fig:as_gauss:walking_L}.

\section{Evaluation of the Human-Aware Navigation Constraint}\label{sec:Simulation-Experiments}
To evaluate the proposed constraints for HAN, two experiments were defined and tested in simulated environments. The proposed system was implemented as an extension of the ROS navigation stack, \cite{article:marder:2010}, and the cost functions, described in the previous section, were implemented as plug-ins to the costmap layered structure, \cite{article:lu:2014}. The simulation environment was Gazebo running on a machine with an Intel Core i7-3770T, and 8GB of RAM.

The experiments performed were:
\newcounter{qcounter}
\begin{list}{\bf Experiment \arabic{qcounter}:\ }
	{\setlength{\labelwidth}{0.5cm}\setlength{\labelsep}{0cm}\setlength{\leftmargin}{0.5cm}\usecounter{qcounter}}
	\item The robot is navigating when encounters a slow walking person, which it must overtake. The goal is to verify if it respects constraints 3 and 6, Sec.~\ref{sec:used-constraints}.
	\item The robot is requested to hand over some object to a person across the room. As it starts moving, the TV is turned on, and the robot replans its path. However, another person is going across the room behind the seated person, who must be overtaken for it to reach its goal. This experiment was designed to evaluate the navigation module regarding constraints 3, 4, 5, and 6, Sec.~\ref{sec:used-constraints}.
\end{list}
The results for these experiments are shown in Fig.~\ref{fig:simulated_environment}.

Next, we present experimental results using both simulated and realistic environments.


\section{Results of the Complete Framework}
\label{sec:results-simul-envir}

In this section, we evaluate the proposed framework (using both the proposed PD and HAN). For that purpose, we use a MBOT mobile platform \cite{72} (see Fig.~\ref{fig:real_setup}\subref{fig:real_setup:mbot}), in a typical indoor work scenario, as shown in Fig.~\ref{fig:real_setup}\subref{fig:real_setup:rviz}. For the PD, we are using five distinct cameras: one onboard and four cameras fixed on the ceiling. Figures showing the images captured by the onboard camera and three of the ceiling cameras are shown in Figs.~\ref{fig:real_setup}\subref{fig:real_setup:camera1}-\subref{fig:real_setup:camera4}.

For the validation, we consider the following experiments:
\begin{list}{\bf Experiment \arabic{qcounter}:\ }
	{\setlength{\labelwidth}{0.5cm}\setlength{\labelsep}{0cm}\setlength{\leftmargin}{0.5cm}\usecounter{qcounter}}
	\item In the first experiments, the robot must avoid three standing people that are distributed along
	the main corridor of the environment, performing a slalom path to avoid entering in their personal space;
	\item The robot is navigating when encounters a slow walking person, which it must overtake. The goal is to verify if it respects constraints 3 and 6, Sec.~\ref{sec:used-constraints}.
	\item A person is seated on a couch, watching TV, and the robot wants to go across the room. The goal is to test if the robot respects constraints 4 and 5, Sec.~\ref{sec:used-constraints}. The TV is an home automated device connected to the same network as the robot, when it is turn on the robot receives a signal through the network.
	\item The robot navigates towards a person, to hand over some object. The goal is to verify the modification of the personal space, constraint 3, Sec.~\ref{sec:used-constraints}.
\end{list}

Notice that, in Experiments 1 and 2 people are detected with Camera 1 to 3 (see Fig.~\ref{fig:real_setup:rviz}). In Experiment 3 we use Camera 4, which is in the testbed \cite{miraldo2015}. Finally, Experiment 4 uses Camera 3. The onboard camera is used in all experiments.

Regarding the HAN throughout the experiments, the robot displayed a similar behaviour to the simulation in terms of trajectory execution. However, the parameters of the cost functions (\ref{eq:f1}), (\ref{eq:f2}), (\ref{eq:f3}), and (\ref{eq:f5}) needed to be adjusted. The values presented previously were derived empirically, taking into account: the values in the literature; the space restrictions of the real scenario, and our intuition of comfort distances.

In the next subsection we discuss the results obtained for each of the experiments.

\subsection{Discussion of the experimental results}

\begin{figure*}
	\centering
	\subfloat[]{\shortstack{\includegraphics[width=0.15\textwidth]{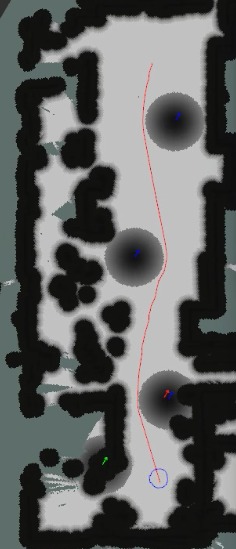}\\ \includegraphics[width=0.15\textwidth]{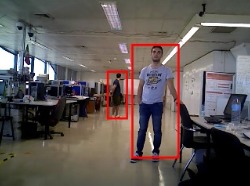}\\ \includegraphics[width=0.15\textwidth]{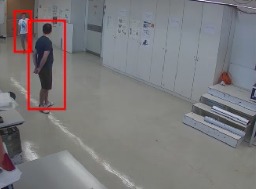}\\ \includegraphics[width=0.15\textwidth]{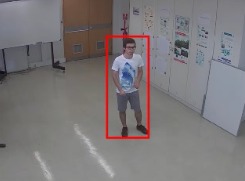}\\ \includegraphics[width=0.15\textwidth]{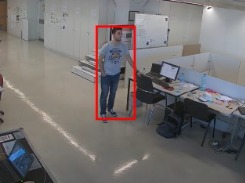}}\label{fig:exp1:1}} \,
	\subfloat[]{\shortstack{\includegraphics[width=0.15\textwidth]{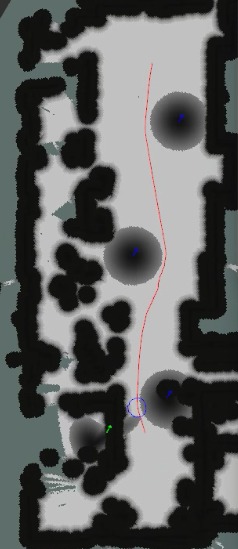}\\ \includegraphics[width=0.15\textwidth]{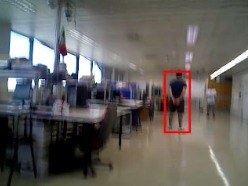}\\ \includegraphics[width=0.15\textwidth]{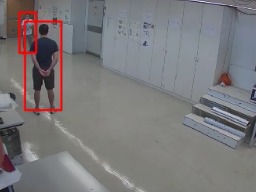}\\ \includegraphics[width=0.15\textwidth]{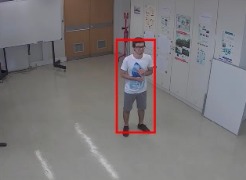}\\ \includegraphics[width=0.15\textwidth]{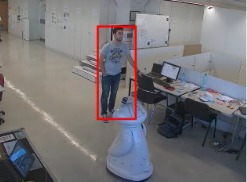}}\label{fig:exp1:2}} \,
	\subfloat[]{\shortstack{\includegraphics[width=0.15\textwidth]{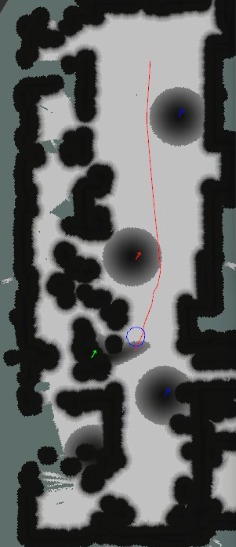}\\ \includegraphics[width=0.15\textwidth]{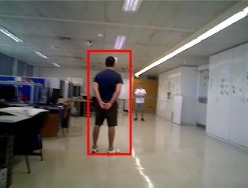}\\ \includegraphics[width=0.15\textwidth]{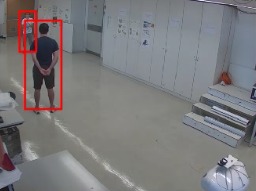}\\ \includegraphics[width=0.15\textwidth]{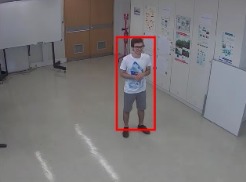}\\ \includegraphics[width=0.15\textwidth]{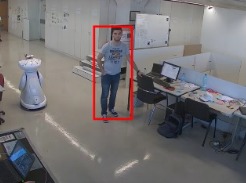}}\label{fig:exp1:3}} \,
	\subfloat[]{\shortstack{\includegraphics[width=0.15\textwidth]{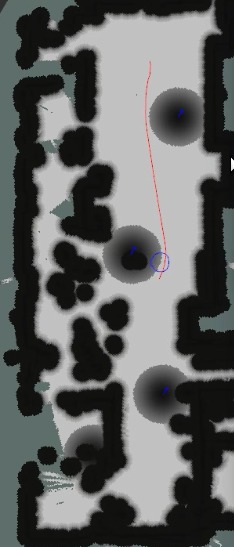}\\ \includegraphics[width=0.15\textwidth]{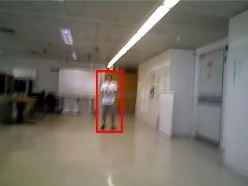}\\ \includegraphics[width=0.15\textwidth]{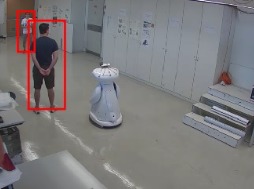}\\ \includegraphics[width=0.15\textwidth]{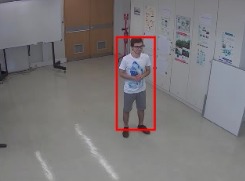}\\ \includegraphics[width=0.15\textwidth]{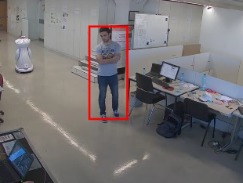}}\label{fig:exp1:4}} \,
	\subfloat[]{\shortstack{\includegraphics[width=0.15\textwidth]{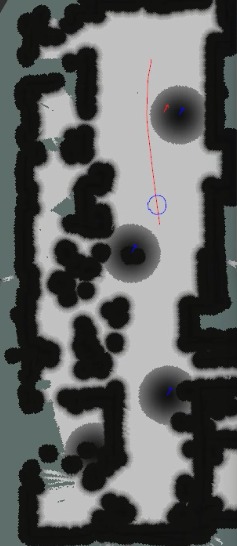}\\ \includegraphics[width=0.15\textwidth]{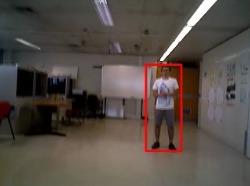}\\ \includegraphics[width=0.15\textwidth]{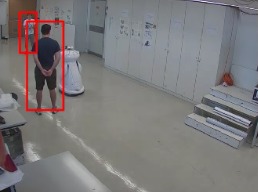}\\ \includegraphics[width=0.15\textwidth]{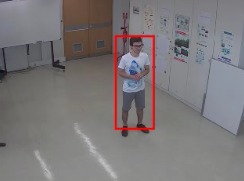}\\ \includegraphics[width=0.15\textwidth]{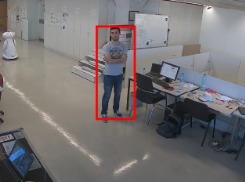}}\label{fig:exp1:5}} \,
	\subfloat[]{\shortstack{\includegraphics[width=0.15\textwidth]{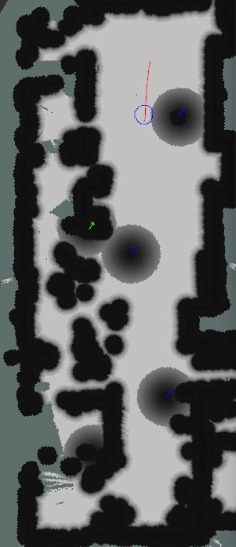}\\ \includegraphics[width=0.15\textwidth]{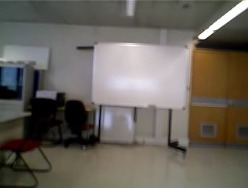}\\ \includegraphics[width=0.15\textwidth]{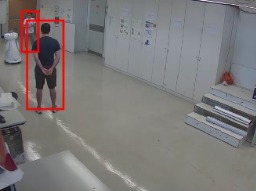}\\ \includegraphics[width=0.15\textwidth]{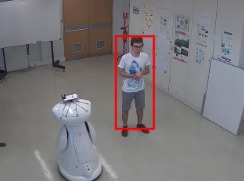}\\ \includegraphics[width=0.15\textwidth]{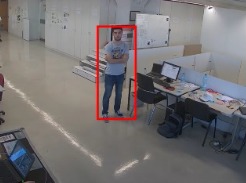}}\label{fig:exp1:6}}
	\caption{In this experiment, people are standing during the robot's movement. From top to bottom, we show: 1) the Rviz representation of the environment, people \& robot positions, and the path planned at each instant; 2) detection from the onboard camera sensor; and 3) three detections from three different cameras mounted on the ceiling. From Figs.~\protect\subref{fig:exp1:1}-\protect\subref{fig:exp1:6} one can see the robot navigation going towards a goal position and avoiding people, while respecting the personal space constraint.}
	\label{fig:exp1}
\end{figure*}

Let us start by the Experiment 1. From the results shown in Fig.~\ref{fig:exp1}, firstly, one can see that the proposed PD correctly detected the pedestrians, including on the onboard sensor, which was able to detect people that were in front of the robot. Using this information, the robot planned and executed the correct path towards the given goal, performing a slalom path to avoid entering in their personal space.

\begin{figure*}
	\centering
	\subfloat[]{\shortstack{\includegraphics[width=0.15\textwidth]{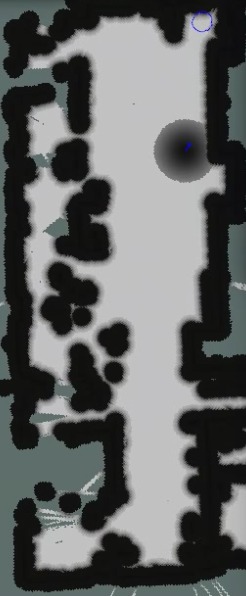}\\ \includegraphics[width=0.15\textwidth]{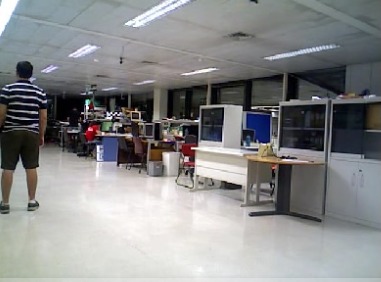}\\ \includegraphics[width=0.15\textwidth]{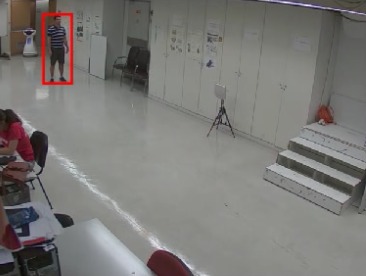}\\ \includegraphics[width=0.15\textwidth]{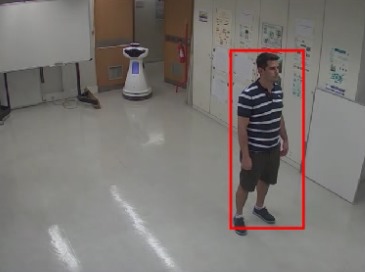}\\ \includegraphics[width=0.15\textwidth]{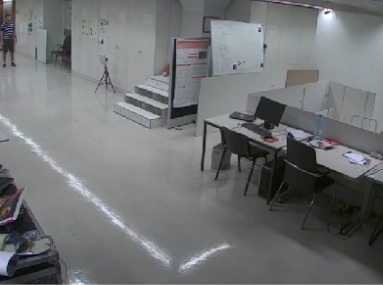}}\label{fig:exp2:1}} \,
	\subfloat[]{\shortstack{\includegraphics[width=0.15\textwidth]{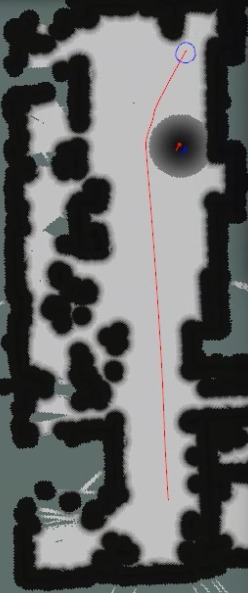}\\ \includegraphics[width=0.15\textwidth]{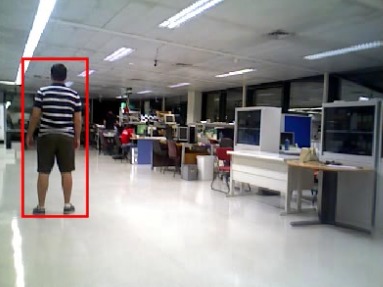}\\ \includegraphics[width=0.15\textwidth]{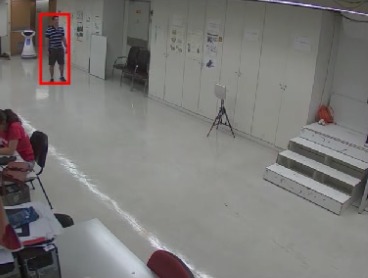}\\ \includegraphics[width=0.15\textwidth]{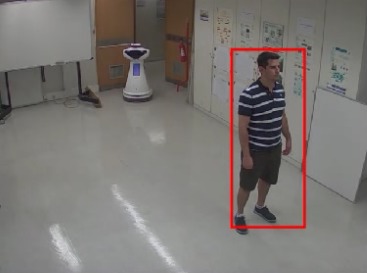}\\ \includegraphics[width=0.15\textwidth]{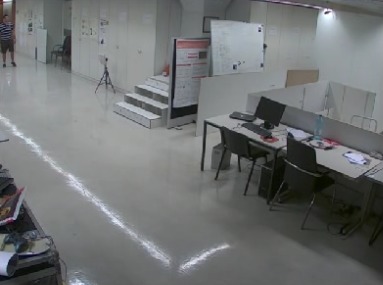}}\label{fig:exp2:2}} \,
	\subfloat[]{\shortstack{\includegraphics[width=0.15\textwidth]{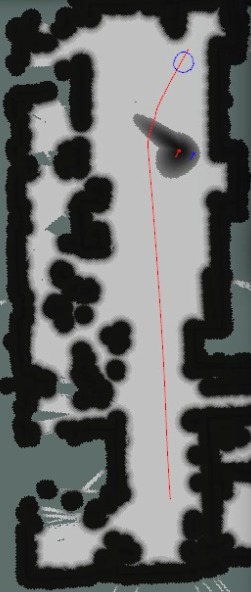}\\ \includegraphics[width=0.15\textwidth]{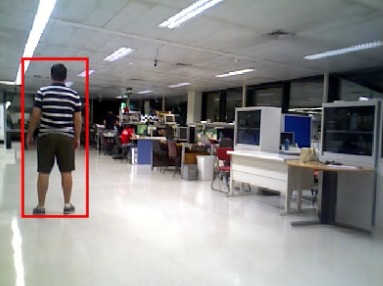}\\ \includegraphics[width=0.15\textwidth]{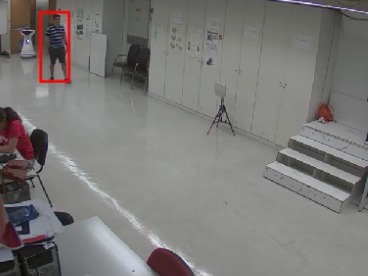}\\ \includegraphics[width=0.15\textwidth]{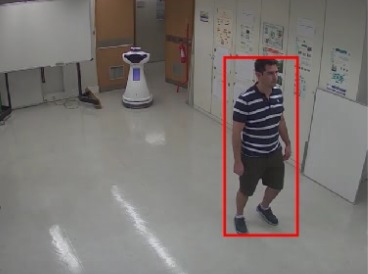}\\ \includegraphics[width=0.15\textwidth]{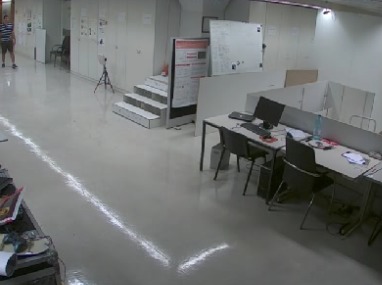}}\label{fig:exp2:3}} \,
	\subfloat[]{\shortstack{\includegraphics[width=0.15\textwidth]{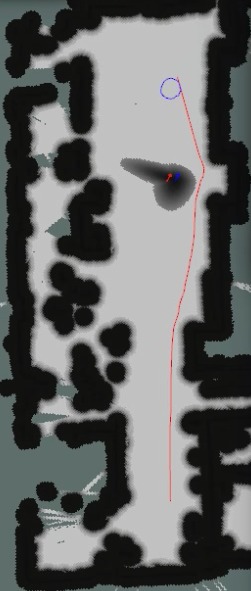}\\ \includegraphics[width=0.15\textwidth]{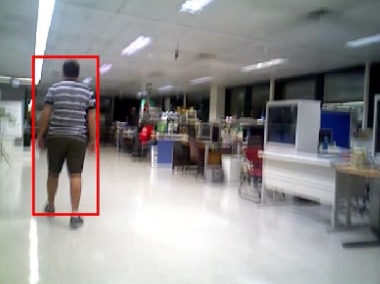}\\ \includegraphics[width=0.15\textwidth]{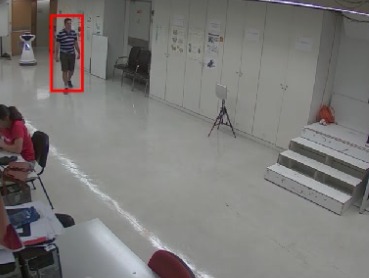}\\ \includegraphics[width=0.15\textwidth]{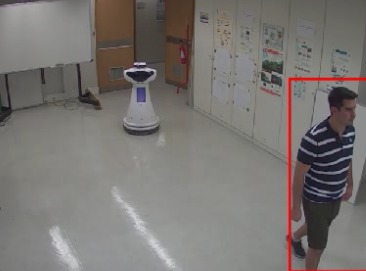}\\ \includegraphics[width=0.15\textwidth]{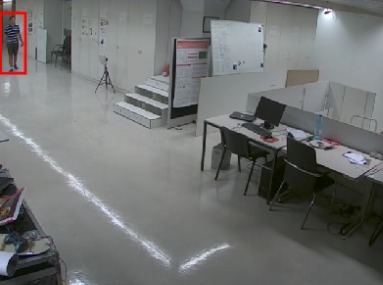}}\label{fig:exp2:4}} \,
	\subfloat[]{\shortstack{\includegraphics[width=0.15\textwidth]{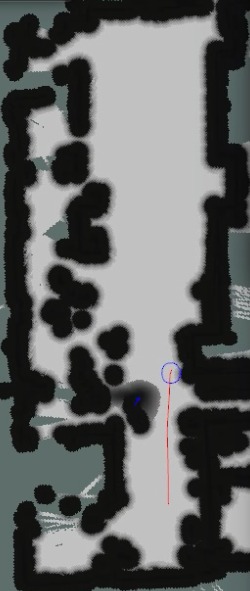}\\ \includegraphics[width=0.15\textwidth]{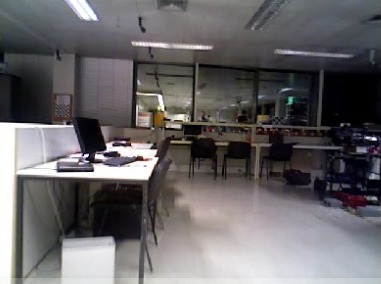}\\ \includegraphics[width=0.15\textwidth]{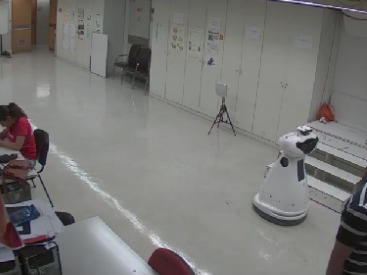}\\ \includegraphics[width=0.15\textwidth]{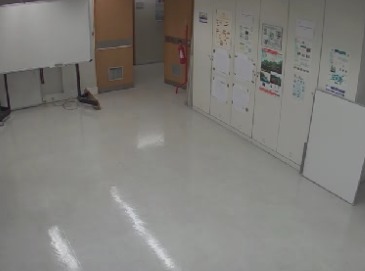}\\ \includegraphics[width=0.15\textwidth]{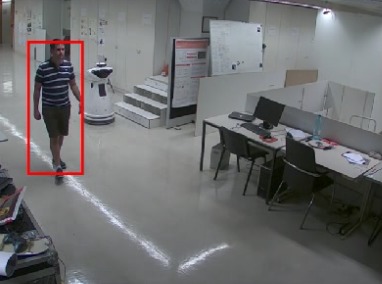}}\label{fig:exp2:5}} \,
	\subfloat[]{\shortstack{\includegraphics[width=0.15\textwidth]{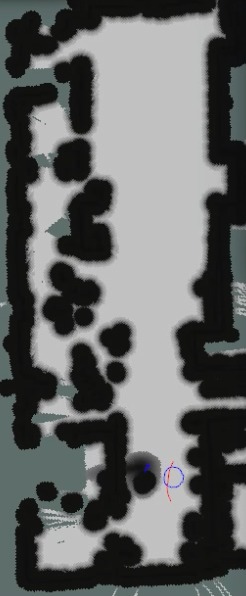}\\ \includegraphics[width=0.15\textwidth]{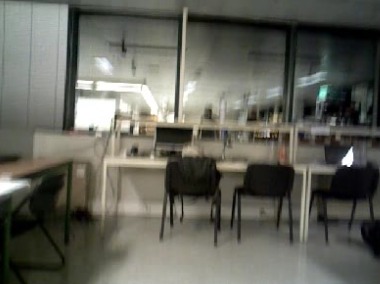}\\ \includegraphics[width=0.15\textwidth]{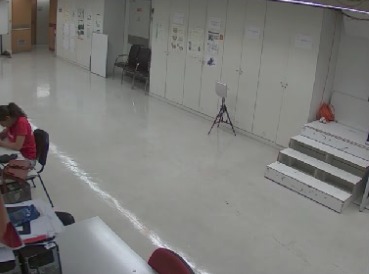}\\ \includegraphics[width=0.15\textwidth]{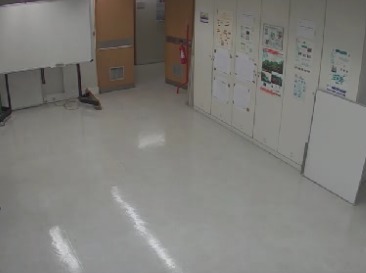}\\ \includegraphics[width=0.15\textwidth]{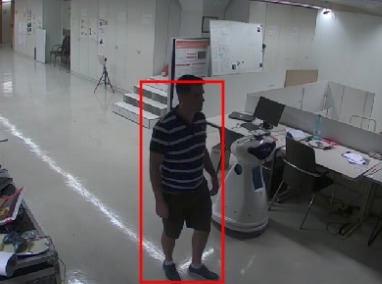}}\label{fig:exp2:6}}
	\caption{Contrarily to the previous experiment, in this case only a person is in the environment and, instead of standing, he starts moving, blocking the robot's path. In this figure we show the environment and images taken from the cameras, similar to Fig.~\ref{fig:exp1}. From Figs.~\protect\subref{fig:exp2:1}-\protect\subref{fig:exp2:6} we can see that firstly the person is identified as standing, but when the robot starts moving, he also starts moving, thus the robot replans its path, in order to overtake the pedestrian by the left. Notice that, since we are using multiple sensors for the PD, even when the person is outside the onboard camera's FOV, the robot performs well because the pedestrian is identified by the external sensors.}
	\label{fig:exp2}
\end{figure*}

Regarding the second experiment, a different setup was considered. Now there is only a person in the environment but, when the robot starts to move, that person will start moving, blocking the robot's path. As it can be seen from Fig.~\ref{fig:exp2}, firstly, the robot plans the path taking into account a person standing in the environment. Then, when the person starts walking the robot replan its path to overtake the pedestrian by the left. The proposed PD was able to detect people from different sensors including the onboard sensor. Since we are using multiple sensors for the detections, even when the onboard camera doesn't see the pedestrian, the robot continues with the right path towards the goal. From these results, one can conclude that the robot was able to correctly detect the pedestrian in real-time and was able to plan its path according with the defined Human-Aware Navigation constraints.

\begin{figure*}
	\centering
	\subfloat[]{\shortstack{\includegraphics[width=0.19\textwidth]{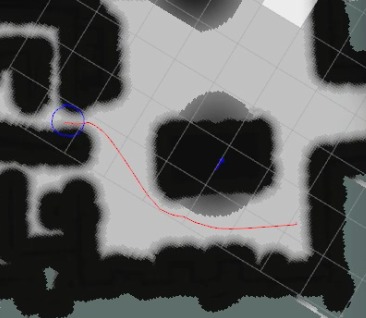}\\\includegraphics[width=0.19\textwidth]{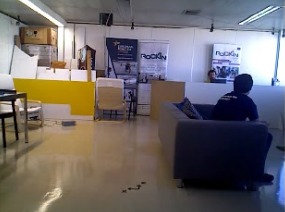}\\\includegraphics[width=0.19\textwidth]{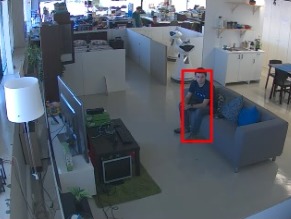}}\label{fig:exp3:1}} \,
	\subfloat[]{\shortstack{\includegraphics[width=0.19\textwidth]{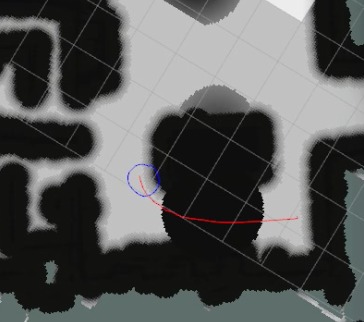}\\\includegraphics[width=0.19\textwidth]{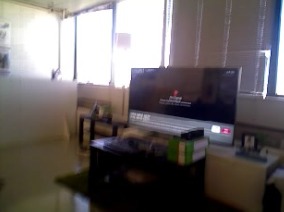}\\\includegraphics[width=0.19\textwidth]{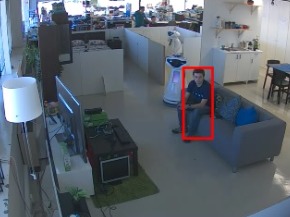}}\label{fig:exp3:2}} \,
	\subfloat[ ]{\shortstack{\includegraphics[width=0.19\textwidth]{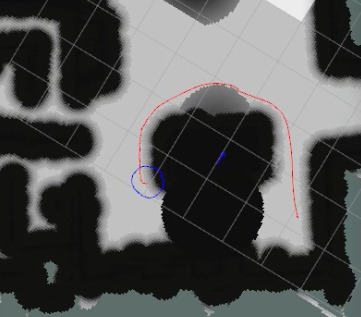}\\\includegraphics[width=0.19\textwidth]{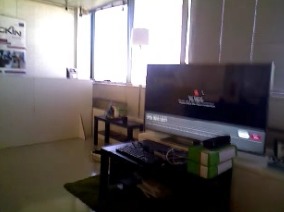}\\\includegraphics[width=0.19\textwidth]{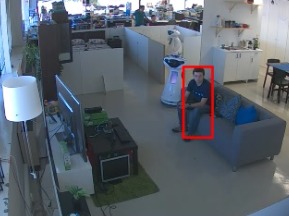}}\label{fig:exp3:3}} \,
	\subfloat[]{\shortstack{\includegraphics[width=0.19\textwidth]{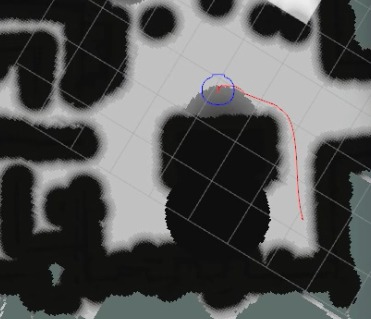}\\\includegraphics[width=0.19\textwidth]{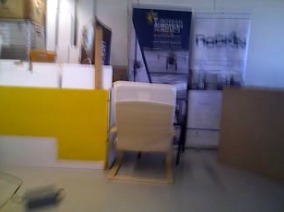}\\\includegraphics[width=0.19\textwidth]{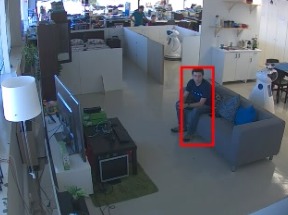}}\label{fig:exp3:4}} \,
	\subfloat[]{\shortstack{\includegraphics[width=0.19\textwidth]{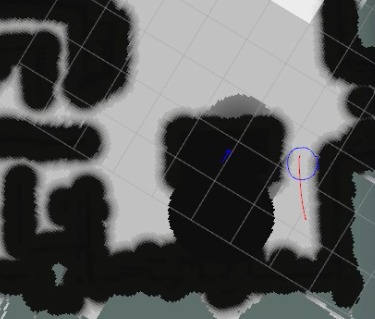}\\\includegraphics[width=0.19\textwidth]{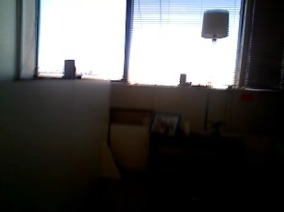}\\\includegraphics[width=0.19\textwidth]{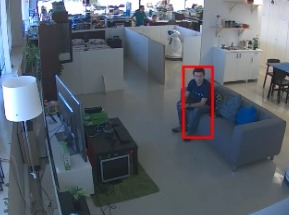}}\label{fig:exp3:5}} \,
	\caption{This figure shows the results for the third experiment. From top to bottom, we show: the representation of the environment (including people \& robot positions and the path planned); the detections from the onboard camera; and the detections from an external camera. Through the results shown in Figs.~\protect\subref{fig:exp3:1}-\protect\subref{fig:exp3:5}, it is possible to observe the two paths planned and followed by the robot when the TV is on and off. Firstly, the robot plans its path passing between the couch and the pedestrian. When the TV is turned on, the robot replans its path to go around the couch.}
	\label{fig:exp3}
\end{figure*}

When considering the third experiment, the goal was to avoid passing in front of a person that was watching TV (a person interacting with an object). These results are shown in Fig.~\ref{fig:exp3}. Firstly, the robot plans its motion according with the less cost path towards the given goal position, which would include passing between the person and the TV. When the robot starts moving, the person turns on the TV and, thus, the robot needs to replan its path, so it does not interfere with the interaction that just started. As can be seen on the second image of Fig.~\ref{fig:exp3}\subref{fig:exp3:1} the pedestrian was partially occluded by the couch on the onboard sensor. Thus the PD running on this sensor was not able to detect the person in the environment. However, since we were using another sensor (external to the robot), which was detecting the person, the robot knew that there was a person there and, when the TV was turned on, the robot replanned the path according with the respective human-aware constraint.

\begin{figure*}
	\centering
	\subfloat[]{\shortstack{\includegraphics[width=0.19\textwidth]{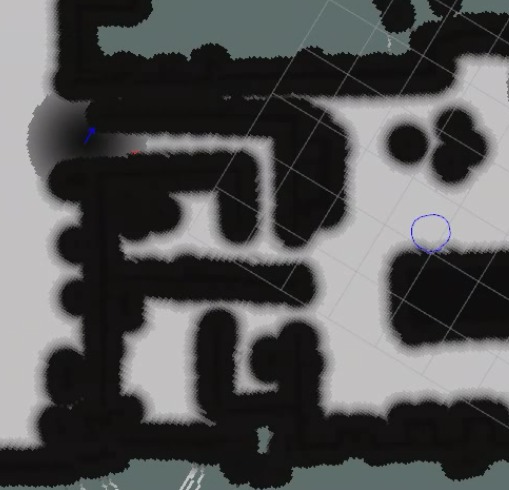}\\\includegraphics[width=0.19\textwidth]{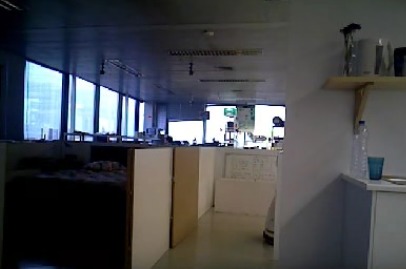}\\\includegraphics[width=0.19\textwidth]{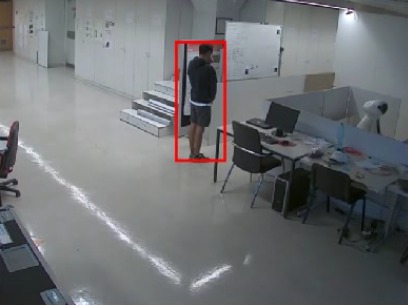}}\label{fig:exp4:1}} \,
	\subfloat[]{\shortstack{\includegraphics[width=0.19\textwidth]{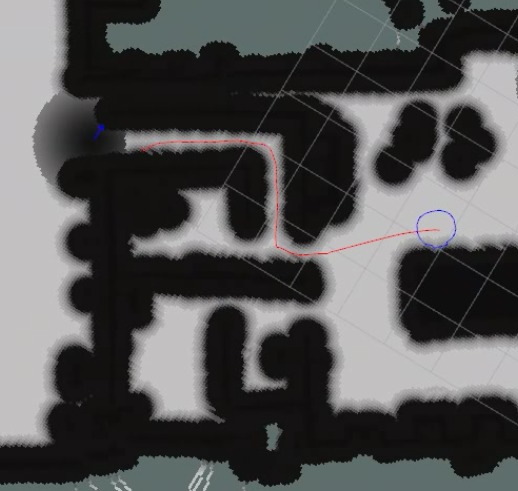}\\\includegraphics[width=0.19\textwidth]{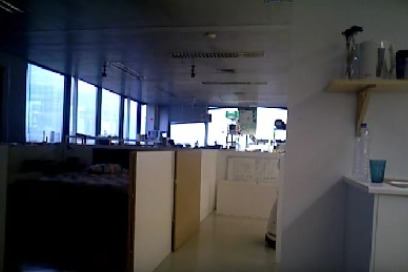}\\\includegraphics[width=0.19\textwidth]{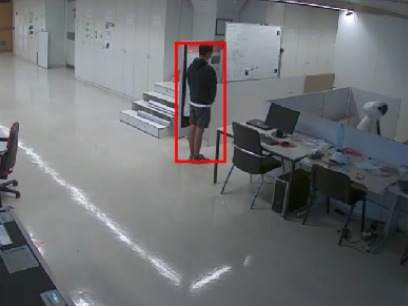}}\label{fig:exp4:2}} \,
	\subfloat[]{\shortstack{\includegraphics[width=0.19\textwidth]{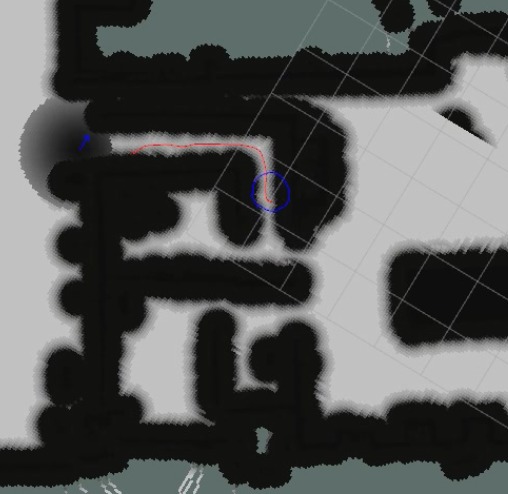}\\\includegraphics[width=0.19\textwidth]{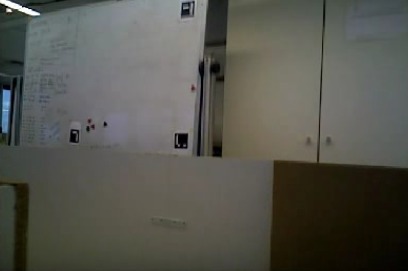}\\\includegraphics[width=0.19\textwidth]{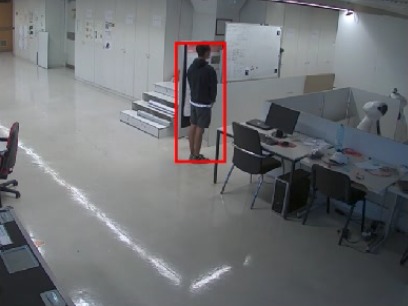}}\label{fig:exp4:3}} \,
	\subfloat[]{\shortstack{\includegraphics[width=0.19\textwidth]{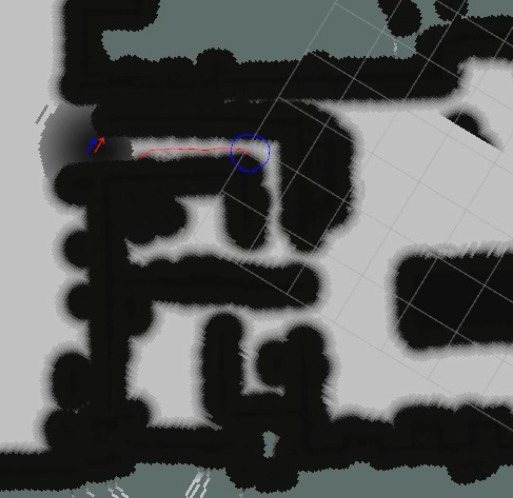}\\\includegraphics[width=0.19\textwidth]{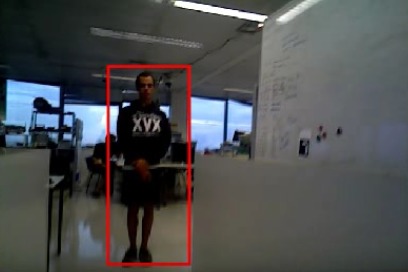}\\\includegraphics[width=0.19\textwidth]{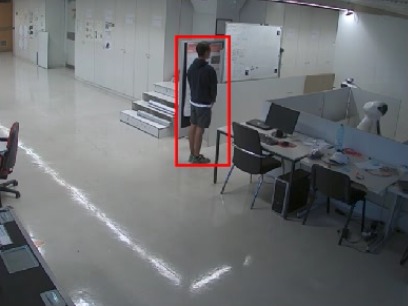}}\label{fig:exp4:4}} \,
	\subfloat[]{\shortstack{\includegraphics[width=0.19\textwidth]{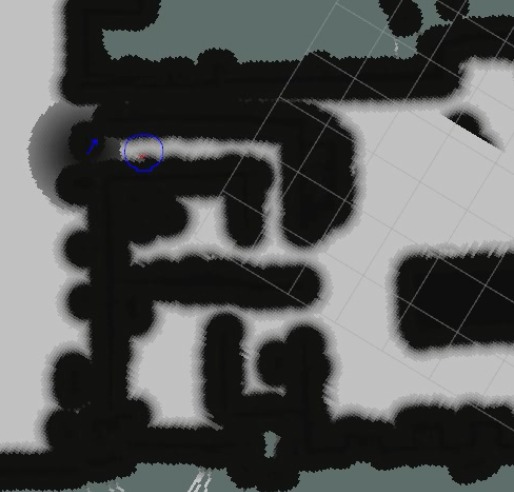}\\\includegraphics[width=0.19\textwidth]{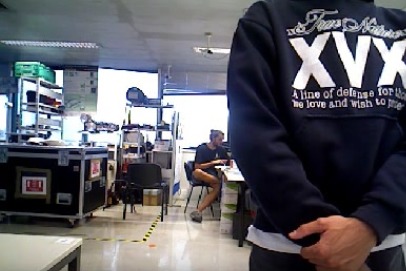}\\\includegraphics[width=0.19\textwidth]{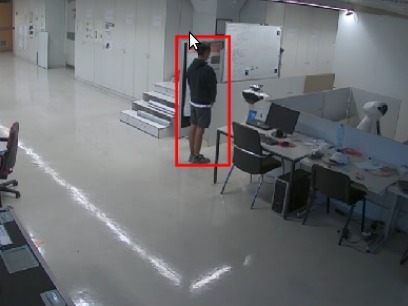}}\label{fig:exp4:5}} \,
	\caption{Similar to the scheme shown in Fig.~\ref{fig:exp3}, from the top to the bottom, we show: the environment representation (with the location of the pedestrian and the robot, as well as the robot's path); detections from the onboard camera; and the detections from an external camera. From Figs:~\protect\subref{fig:exp4:1}-\protect\subref{fig:exp4:5}, it is possible to observe that the robot follows the path planned to the hand over position, and that the costmap of the personal space is changed so the robot can reach that position.}
	\label{fig:exp4}
\end{figure*}

In the last experiment, we show a case where the robot is asked to receive some object from a person. In this case, the robot should identify the person's position, which is given by the proposed PD algorithm. Then, the robot should navigate towards his position and enter in his personal space to be able to receive the object. As it can be seen from Fig.~\ref{fig:exp4}, both the proposed PD and the HAN work as expected, even in the case where the pedestrian is not seen by the onboard sensor.

A video with these experiments is sent as supplementary material and ROS packages for both the PD and the HAN will be available to the community.


\section{Conclusions}
\label{sec:conclusions}

This work addresses the problem of Human-Aware Navigation for a robot in a social context. For that purpose, in this paper we derive a robust and efficient solution for the PD, using deep learning. In addition, regarding HAN, we reformulate some of the respective constraints in order to have a standardization of human-aware constraints.

To validate our contributions, we first use the INRIA dataset to evaluate both the accuracy and runtime figures of the PD method. Then, to evaluate the performance using real data, we use two sequences of images (one was acquired using the robot's onboard camera and the second sequence was acquired using an external camera). The results show that the method is both robust and fast enough (up to 10 frames per second) for robot navigation applications. To evaluate the standardization of HAN constraints, we use a simulated environment.

In a realistic scenario, we test both modules in four distinct scenarios. The results show that the robot has the correct behavior, which means that both PD and HAN are working properly.

\section*{Acknowledgements}
This work was partially supported by the Portuguese Foundation for Science and Technology (FCT) grants {\tt PD/BD/135015/2017} (through the NETSys Doctoral Program) \& {\tt SFRH/BPD/111495/2015}, and ISR/LARSyS Strategic Funding by the FCT project  {\tt PEst-OE/EEI/LA0009/2013}.

\section*{References}

\bibliographystyle{elsarticle-num}
\bibliography{./IEEEabrv,root}

\appendix

\section{CNN model}
\label{sec:CNN-model}

We now formalize the main ingredients of the CNN architecture.
Basically, this type of deep networks comprises several processing
stages. Each stage is characterized by having two types of layers,
namely: a convolutional layer containing a non-linear activation
function, and a non-linear subsampling layer. In the former, a
convolutional filter is applied to the input. In the latter, the
size reduction of the input is achieved. These two stages are
typically followed by several fully connected layers, and a
multinomial logistic regression layer (see details in
\cite{KrizhevskyNIPS2012}). Formally, the convolutional neural
network can be analytically represented by the following mapping
$f:{\cal X}\rightarrow {\cal Y}$, where ${\cal X}$ represents the
image space and ${\cal Y}$ represents the classification space:
\begin{equation}
f({\bf v}; \theta) = {\bf v}^{\star}=
f_{\rm out}\circ f_{\rm f_{c}} \circ s_{L} \circ a_L \circ c_L \circ ... \circ s_1 \circ a_1 \circ c_1({\bf v}^{(0)}),\label{eq:CNN1}
\end{equation}
where $\circ$ denotes the composition operator, $\{c_i(.)\}_{i=1}^L$ represents a \texttt{convolutional} layer, $\theta$ represents the model parameters
comprising the input weight matrices ${\bf W}_l\in\mathbb{R}^{k_l\times k_l\times n_l\times n_{l-1}}$ and bias vector $\beta_l\in\mathbb{R}^{n_l}$  for each layer
$l\in\{1,...,L\}$, and with $k_l\times k_l$ representing the size of the $n_l$ filters in the $l$-th layer, having $n_{l-1}$ input channels ;
$a_{l}(.)$ represents a non-linear \texttt{activation} layer (e.g. the Rectified Linear Unit (ReLU), for more details see \cite{KrizhevskyNIPS2012}) ; $s_{l}(.)$ is a \texttt{sub-sampling} layer, that is, a function that allows to obtain ${\bf v}^{l}= \downarrow {\bf v}^{(l-1)}$, where $\downarrow$ denotes a subsampling function that pools (using the mean or max functions) the values from a region of the input data ; $f_{\rm f_{c}}$ is a \texttt{fully-connected} layer containing the weights $\{{\bf W}_{f_{\rm c},k}\}_{k=1}^K$ (with  ${\bf W}_{f_{\rm c},k}\in\mathbb{R}^{n_{{f_{\rm c}},k-1}\times n_{{f_{\rm c}},k}}$ representing the connections between the $k-1$-th and $k$-th fully connected layers), and biases  $\{{\beta}_{f_{\rm c},k}\}_{k=1}^K$
(with $\beta\in\mathbb{R}^{n_{f_{\rm c},k}}$), that also belong to the model parameters $\theta$; $f_{\rm out}$ is a \texttt{multinomial logistic regression} layer containing the weights ${\bf W}_{\rm out}\in\mathbb{R}^{n_{{f_{\rm c}},K}\times C}$ and bias $\beta_{\rm out}\in\mathbb{R}^{C}$ ($C$ is the number of classes under consideration).

The output of the CNN mentioned in (\ref{eq:CNN1}), can be seen as an approximation of the input data (represented by
${\bf v}^{\star}$ in equation (\ref{eq:CNN1})). The convolution mentioned above is formally defined as:
\begin{equation}
c_{l}({\bf v}^{(l-1)}(j)) = \sum_{i\in\Omega(j)} {\bf v}^{(l-1)}(i) \star {\bf W}^{l}(i,j) + \beta^{l}(j),\label{eq:CNN2}
\end{equation}
where $\star$ stands for the convolution, $\Omega(j)$ is the input region addresses and where the convolutional filters are represented by the weight matrix ${\bf W}^l$ and the bias vector $\beta^l$.
Notice that the input ${\bf v}^{(l-1)}(j)$ in (\ref{eq:CNN2}), is obtained following the structure in (\ref{eq:CNN1}), i.e., convolution, activation and sub-sampling operations \footnote{These are the basic operations of the CNN, but the subsampling layer does not necessarily need to be present in every case.}, from the preceding layer, that is:
\begin{equation}
{\bf v}^{(l-1)}(j) = s_{l-1}\Bigl(a_{l-1}(c_{l-1}({\bf v}^{(l-2)}(j) ))\Bigr),
\end{equation}
where ${\bf v}^{(0)}(j)$  represents the input image.

The $L$ convolutional layers are followed by a sequence of fully
connected layers, that perform a particular instance of the
convolution in (\ref{eq:CNN2}) to the entire vectorised input ${\bf
	v}^L\in\mathbb{R}^{|{\bf v}^L|}$, where $|{\bf v}^L|$ denotes the
length-vector ${\bf v}^L$. In the final stage, the above fully
connected layers are followed by a classification layer that is
defined by a soft-max function as follows (see
\cite{KrizhevskyNIPS2012}):
\begin{equation}
f_{\rm out}(f_{{f_c}}) = {\rm softmax}({\bf W}_{\rm out} f_{{f_c}} + \beta_{\rm out}), \label{eq:out}
\end{equation}
with the soft-max function defined as  ${\bf y}({\bf
	q})_i=\frac{\exp({{\bf q}}(i))}{\sum_j \exp({{\bf q}}(j))}$ and
$f_{\rm out}\in[0,1]^C$ represents the output from the inference
process that takes the input ${\bf v}$, with $C$ representing the
number of classes.

In our case, the input consists of the proposals of the \texttt{ACF} non-deep detector, i.e., ${\bf x}({\cal B})$, and $C$ represents the two output classes
(absence/presence of the pedestrian). Thus, \eqref{eq:CNN1} is written as (similarly for \eqref{eq:CNN2}):
\begin{equation}
\begin{split}
f({\bf x}&({\cal B});\theta) = {\bf x}({\cal B })^{\star}\\
&=f_{\rm out}\circ f_{\rm f_{c}} \circ s_{L} \circ a_L \circ c_L\circ ... \circ s_1 \circ a_1 \circ c_1({\bf x}({\cal B}^{(0)})),
\end{split}
\end{equation}
where the inputs are the proposals (i.e., the image content, in the RGB feature map, delimited by the bounding boxes), here denoted as ${{\bf x}(\cal B}^{(0)})$ (see Fig. \ref{fig:PD-proposal}). The main idea is to take the proposals ${\bf x}({\cal B})$, that will be processed by the CNN, and produce a classification probability that a given proposal contains a pedestrian. The proposals classified as non pedestrians are discarded, allowing to eliminate false positives. The ones regarded as pedestrians are kept, including the original ACF detector score.

In the PD case, the CNN prediction output can be formally represented by:
\begin{equation}
f({\bf x}({\cal B}),\theta)= y^{\star},
\end{equation}
which is trained using the binary cross-entropy loss over the training set indexed by $i$, as follows:
\begin{equation}
{\cal L} = \frac{1}{|{\cal D}|}\sum_{i=1}^{|\cal D|} - y(i)\times \log(y^{\star}(i)) - (1-y(i)) \times \log(1-y^{\star}(i))\label{eq:binary-loss}
\end{equation}

Let the pre-trained CNN be represented by the model $\widetilde{\bf y}=f(\widetilde{\bf x},\widetilde{\theta})$, with $\widetilde\theta=[\widetilde\theta_{\rm cn},\widetilde\theta_{\rm fc},\widetilde\theta_{\rm lr}]$. The process of pre-training a CNN  is defined by the following three steps:
\begin{enumerate}
	\item{Training $M_1$ stages of convolutional and non-linear subsampling layers, that are represented by the parameters $\widetilde{\theta}_{\rm cn}$; then}
	\item{Training $M_2$ fully connected layers, represented by the parameters $\widetilde{\theta}_{\rm fc}$; and}
	\item{Training one multinomial logistic regression layer with parameters $\widetilde{\theta}_{\rm lr}$, by minimizing the cross-entropy loss function \cite{KrizhevskyNIPS2012} over the dataset $\widetilde{\cal D}$.}
\end{enumerate}

It is worth mentioning that, transferring a large number of
pre-trained layers and fine-tuning the CNN is the key to achieve the
best classification results in transfer learning problems
\cite{YosinskiNIPS2014}. Following this strategy, we first take the
$M_1$ layers to initialize a new model (see
\cite{YosinskiNIPS2014}). Since we have changed the CNN input size
to reduce the computational expense, the $M_2$ layers were  randomly
initialized from a Gaussian distribution, so that the dimensions are
compatible and inference is possible. Finally, we introduce a new
binomial logistic regression layer, with parameters ${\theta}_{\rm
	lr}$ (randomly initialized from a Gaussian distribution) adapted for
two classes (pedestrian and non-pedestrian). Afterwards, we
fine-tune all the parameters in all the layers of the CNN model (i.e., $M_1$, $M_2$ and the multinomial logistic regression layer) by minimizing the cross-entropy loss
function in (\ref{eq:binary-loss}), using the pedestrian training
set ${\cal D}$.


\section{Runtime comparison with other related approaches}
\label{sec:comp_cascades_exhaustive_search}

In this section an analysis is conducted for the exhaustive search CNN classification procedure, in order to compare its speed with the one obtained with the proposed method (i.e., the cascade ACF+CNN, including a threshold operation). We also justify the choice in the use of proposed cascade, instead of using the faster R-CNN \cite{{ren2015faster}}.

The number of CNN classification iterations for the exhaustive search procedure, using a single scale (instead of multiscale) are: $N_h=\lfloor{(H-h)/s} \rfloor +1$, for the height; and $N_w=\lfloor{(W-w)/s} \rfloor +1$, for the width; where $(H,W)$ denote the height and width of the entire image (e.g. $480 \times 640$), respectively; $(h,w)$ represent the detection windows height and width (e.g. $64 \times 64$), respectively; s denotes the stride (e.g. 4 pixels).

The total exhaustive search CNN classification time per image $T_{exh}$ is: $T_{exh}= (N_h + N_w)\cdot t_{CNN}$, where $t_{CNN}$ is the time spent in a CNN feedforward computation.

If the image dimensions correspond to the ones used in the evaluation on real scenarios (in Section \ref{sec:Results-CNN-Corridor-MBOT}), the detection window size corresponds to the adapted VGG-VD16 model input dimensions (mentioned in Section \ref{sec:Adaptation-of-the-deep-network}), the stride follows the value from \cite{DollarPAMI2014}, and the CNN classification time corresponds to the one from the adapted VGG-VD16 model (and fine-tuned for PD, as described in Section \ref{sec:Adaptation-of-the-deep-network}), then $(H,W) = (480,640)$; $(h,w) = (64,64)$; $s=4$; $N_h = \lfloor{(480-64)/4}\rfloor +1 = 105$ iterations, and $N_w=\lfloor{(640-64)/4}\rfloor +1 = 145$ iterations; $t_{CNN}=0.0323$ seconds. Therefore, $T_{exh}= (N_h + N_w) \cdot t_{CNN} = (105+145) \cdot 0.0323 = 8.075$ seconds, or $0.1238$ FPS.

Consequently, we conclude that the exhaustive search procedure is not adequate for HAN tasks, because of its runtime, which is $0.1238$ FPS (using only a single scale). Furthermore, the proposed approach, comprising a cascade of ACF and CNN (including a threshold operation), is substantially faster than the exhaustive search (reaching approximately 10 FPS, as mentioned in Table \ref{tab:running-times-2}), while using multiple scales.

The use of the faster R-CNN can also be adopted. Notice that the faster R-CNN can be fully explored in GPU based implementations. However, such requirement can not be used in our experimental setup. More specifically, our robotic setup is composed of two main parts: body and head. The head can pan and has LED backlight to express emotions through a drawn mouth, eyes and checks. The body has all the CPU devices (two motherboards with i7 processors), a touch-screen and all of the navigation mechanics, based on a Four-Wheel Omnidirectional Mecanum drive. Therefore, we opted to use an architecture that allows to efficiently detect people resorting to a CPU instead of employing GPU. Furthermore, since the method is CPU based, it can be easily implemented in generic robotic settings (following the specifications mentioned above).

\end{document}